\documentclass[conference]{IEEEtran}
\usepackage{times}

\IEEEoverridecommandlockouts 
\usepackage[numbers, sort&compress]{natbib}

\usepackage{multicol}
\usepackage[bookmarks=true]{hyperref}

\usepackage{algorithm}
\usepackage{algorithmic}

\usepackage{graphics} 
\usepackage{epsfig} 
\usepackage{mathrsfs}
\usepackage{times} 
\usepackage{amsmath} 

\usepackage{amsthm}
\usepackage{amssymb}  
\usepackage{gensymb}

\usepackage[footnotesize]{caption} 

\usepackage{float,bm}
\usepackage{wasysym}
\usepackage[normalem]{ulem}
\usepackage{mathtools}
\usepackage{bbm}
\usepackage{graphicx}
\usepackage{caption}
\usepackage{subcaption}
\usepackage{color}  
\usepackage[dvipsnames]{xcolor}
\usepackage{titlesec}
\usepackage{indentfirst}
\usepackage{multirow}

\usepackage{perl_acronyms}
\usepackage{perl_misc}

\usepackage{color, colortbl}

\usepackage{soul} 

\begin{document}

\title{Phasing Through the Flames: Rapid Motion Planning with the AGHF PDE for Arbitrary Objective Functions and Constraints}
\author{Challen Enninful Adu$^{*}$, César E. Ramos Chuquiure$^{*}$, Yutong Zhou, Pearl Lin, Ruikai Yang, Bohao Zhang,\\ Shubham Singh and Ram Vasudevan }
\newtheorem{defn}{Definition}
\newtheorem{rem}[defn]{Remark}
\newtheorem{lem}[defn]{Lemma}
\newtheorem{prop}[defn]{Proposition}
\newtheorem{assum}[defn]{Assumption}
\newtheorem{ex}[defn]{Example}
\newtheorem{runx}{Running Example}
\newtheorem{thm}[defn]{Theorem}
\newtheorem{cor}[defn]{Corollary}
\newtheorem{problem}[defn]{Problem}

\definecolor{init_pose_turq}{HTML}{3FC8D1}
\definecolor{final_pose_gold}{HTML}{D8A621}
\definecolor{obs_red}{HTML}{C80000}
\definecolor{init_traj_maroon}{HTML}{BD3F00}
\definecolor{phase1_traj_mustard}{HTML}{C9B305}
\definecolor{blaze_color}{HTML}{349C77}
\definecolor{raptor_color}{HTML}{7770B2}
\definecolor{aligator_color}{HTML}{A27725}
\definecolor{init_traj_orange}{HTML}{F46700}
\definecolor{phase1_traj_blue}{HTML}{0A007F}
\definecolor{final_traj_green}{HTML}{329400}

\newcommand{\comment}[1]{\textnormal{{\textcolor{gray}{#1}}}}
\providecommand{\Ram}[1]{{\textnormal{\color{WildStrawberry}\textbf{Ram: #1}}}}
\providecommand{\ram}[1]{\Ram{#1}}
\newcommand{\challen}[1]{{\textnormal{\color{BlueGreen}\textbf{Challen: #1}}}}
\newcommand{\bohao}[1]{{\textnormal{\color{Red}\textbf{Bohao: #1}}}}

\newcommand{\cesar}[1]{{\textnormal{\color{Cyan}\textbf{(Cesar: #1)}}}}

\providecommand{\methodname}{\text{PHLAME}}
\providecommand{\methodnamenew}{\text{BLAZE}}

\newcommand{\new}[1]{{\color{Plum}{#1}}}
\providecommand{\resp}[1]{\new{#1}}
\newcommand{\old}[1]{{\color{Yellow}{(old) #1}}}

\newcommand{\strikethrough}[1]{{\color{Orange}{\st{#1}}}}

\newcommand{\stkout}[1]{{\color{Orange}\ifmmode\text{\sout{\ensuremath{#1}}}\else\sout{#1}\fi}}

\providecommand{\Int}{\texttt{int}}
\providecommand{\INT}{\texttt{INT}}
\providecommand{\Sum}{\texttt{sum}}

\newcommand{\algorithmicbreak}{\textbf{break}}
\newcommand{\BREAK}{\STATE \algorithmicbreak}
\providecommand{\PP}{\mathcal{P}}
\providecommand{\LL}{\mathcal{L}}
\providecommand{\W}{\mathcal{W}}
\providecommand{\D}{\mathcal{D}}
\providecommand{\X}{x}
\providecommand{\Xs}{x_s}
\providecommand{\Xstar}{\mathcal{X}^{*}}
\providecommand{\K}{\mathcal{K}}
\providecommand{\Z}{\mathcal{Z}}
\providecommand{\XY}{\mathcal{XY}}
\renewcommand{\P}{\mathcal{P}}
\providecommand{\G}{\mathcal{G}}
\providecommand{\B}{\mathcal{B}}
\providecommand{\Z}{\mathcal{Z}}
\providecommand{\A}{\mathcal{A}}
\providecommand{\V}{\mathcal{V}}
\providecommand{\U}{\mathcal{U}}
\providecommand{\T}{\mathcal{T}}
\providecommand{\Y}{\mathcal{Y}}
\providecommand{\RR}{\mathcal{R}}
\providecommand{\Q}{\mathcal{Q}}
\providecommand{\HH}{\mathcal{H}}
\providecommand{\I}{\mathcal{I}}
\providecommand{\M}{\mathcal{M}}
\providecommand{\J}{\mathcal{J}}
\providecommand{\E}{\mathcal{E}}
\providecommand{\F}{\mathcal{F}}
\renewcommand{\SS}{\mathcal{S}}
\providecommand{\W}{\mathcal{W}}
\providecommand{\OO}{\mathcal{O}}
\providecommand{\R}{\ensuremath \mathbb{R}}
\providecommand{\N}{\ensuremath \mathbb{N}}

\providecommand{\psX}{\xi}
\providecommand{\psx}{\xi}
\providecommand{\Jp}{J_{\Xi}}
\providecommand{\Jpi}{J_{\Xi_i}}

\providecommand{\Gx}{G(X)}
\providecommand{\G}{G}
\providecommand{\uu}{u}
\providecommand{\x}{\mathtt{x}}
\providecommand{\xd}{\dot{\mathtt{x}} }
\providecommand{\xdd}{\ddot{\mathtt{x}} }
\providecommand{\Xd}{\dot \X}
\providecommand{\Xdd}{\ddot \X}
\providecommand{\Fd}{F_d}
\providecommand{\F}{F}
\providecommand{\Fbar}{\bar{F}}
\providecommand{\Fbarx}{\bar{F}(\x(t))}
\providecommand{\Fc}{F_c}
\providecommand{\Fcx}{F_c(\x(t))}
\providecommand{\Fdx}{F_d(\x(t))}
\providecommand{\Fx}{F(\x(t))}

\providecommand{\q}{q}
\providecommand{\qd}{\dot{q}}
\providecommand{\qdd}{\ddot{q}}
\providecommand{\Hq}{H(q(t))}
\providecommand{\Hinvq}{H^{-1}(\x_{P1}(t))}
\providecommand{\Hinv}{H^{-1}}
\providecommand{\Cq}{C(\x_{P1}(t),\x_{P2}(t))}
\providecommand{\G}{G}
\providecommand{\Ginv}{G^{-1}}
\providecommand{\M}{M}
\providecommand{\Minv}{M^{-1}}
\providecommand{\dGdx}{\frac{\partial G}{\partial \X}}
\providecommand{\dFddx}{\frac{\partial \Fd}{\partial \X}}
\providecommand{\dFdTdx}{\frac{\partial \Fd^T}{\partial \X}}

\providecommand{\Lxxd}{L(\X(t),\Xd(t))}
\providecommand{\Lxsxds}{L(\X_s(t),\Xd_s(t))}
\providecommand{\Lxxds}{L(\X(t,s),\Xd(t,s))}
\providecommand{\Act}{\mathcal{A}}
\providecommand{\dxds}{\frac{\partial \X}{\partial s}}
\providecommand{\dxsds}{\frac{\partial \X_s}{\partial s}}
\providecommand{\dxdsds}{\frac{\partial \Xd_s}{\partial s}}
\providecommand{\dds}{\frac{\partial}{\partial s}}
\providecommand{\ddt}{\frac{d}{dt}}
\providecommand{\dldx}{\frac{\partial L}{\partial \X}}
\providecommand{\dldxd}{\frac{\partial L}{\partial \dot{\X}}}
\providecommand{\dldxs}{\frac{\partial L}{\partial \X_s}}
\providecommand{\dldxds}{\frac{\partial L}{\partial \dot{\X}_s}}

\providecommand{\dludx}{\frac{\partial L_u}{\partial \X}}
\providecommand{\dludxd}{\frac{\partial L_u}{\partial \dot{\X}}}
\providecommand{\dludxs}{\frac{\partial L_u}{\partial \X_s}}
\providecommand{\dludxds}{\frac{\partial L_u}{\partial \dot{\X}_s}}

\providecommand{\dlddx}{\frac{\partial L_d}{\partial \X}}
\providecommand{\dlddxd}{\frac{\partial L_d}{\partial \dot{\X}}}
\providecommand{\dlddxs}{\frac{\partial L_d}{\partial \X_s}}
\providecommand{\dlddxds}{\frac{\partial L_d}{\partial \dot{\X}_s}}

\providecommand{\dldxt}{\frac{\partial L}{\partial \X_s(t)}}
\providecommand{\dldxdt}{\frac{\partial L}{\partial \dot{\X}_s(t)}}
\providecommand{\dldxtx}{\frac{\partial L(\X_s(t),\dot{\X}_s(t)) }{\partial \X_s(t)}}
\providecommand{\dldxdtx}{\frac{\partial L(\X_s(t),\dot{\X}_s(t)) }{\partial \dot{\X}_s(t)}}
\providecommand{\xstar}{\x^{*}}
\providecommand{\xdstar}{\xd^{*}}
\providecommand{\xtstar}{\x^{*}(t)}
\providecommand{\xstardot}{\x^{*}(\cdot)}
\providecommand{\smax}{s_{max}}
\providecommand{\xtsmax}{\x(t,s_{max})}
\providecommand{\xdtsmax}{\xd(t,s_{max})}
\providecommand{\xdts}{\xd(t,s)}

\providecommand{\xpone}{\X_{P1}}
\providecommand{\xptwo}{\X_{P2}}
\providecommand{\xponest}{\X_{P1}(s,t)}
\providecommand{\xptwost}{\X_{P2}(s,t)}
\providecommand{\xdpone}{{\dot \X}_{P1}}
\providecommand{\xdptwo}{{\dot \X}_{P2}}
\providecommand{\xdponest}{{\dot \X}_{P1}(s,t)}
\providecommand{\xdptwost}{{\dot \X}_{P2}(s,t)}
\providecommand{\xddpone}{{\ddot \X}_{P1}}
\providecommand{\xddptwo}{{\ddot \X}_{P2}}

\providecommand{\Hdot}{{\dot H}}
\providecommand{\HdotT}{{\dot H^T}}
\providecommand{\HT}{{H^T}}
\providecommand{\Cdot}{{\dot C}}
\providecommand{\h}{{\mathcal{h}}}

\providecommand{\dXdt}{{\frac{\partial \x}{\partial t}}}
\providecommand{\ddXdt}{{\frac{\partial^2 \x}{\partial t^2}}}
\providecommand{\dddXdt}{{\frac{\partial^3 \x}{\partial t^3}}}

\providecommand{\FDuzero}{FD_{0}}

\providecommand{\dHinvCdxpone}{ \frac{\partial}{\partial \xpone}\big(\Hinv C\big) }
\providecommand{\dHinvCdxptwo}{ \frac{\partial}{\partial \xptwo}\big(\Hinv C\big) }
\providecommand{\dmHinvCdxpone}{ \frac{\partial \FDuzero}{\partial \xpone} }
\providecommand{\dmHinvCdxptwo}{ \frac{\partial \FDuzero}{\partial\xptwo} }
\providecommand{\dFDdxpone}{ \frac{\partial \FDuzero}{\partial \xpone} }
\providecommand{\dFDdxptwo}{ \frac{\partial \FDuzero}{\partial \xptwo} }
\providecommand{\dFDTdxpone}{ \frac{\partial \FDuzero^T}{\partial \xpone} }
\providecommand{\dFDTdxptwo}{ \frac{\partial \FDuzero^T}{\partial\xptwo} }

\providecommand{\ddFDddxpone}{\frac{\partial^2 FD_0}{\partial \xpone^2}}
\providecommand{\ddFDddxptwo}{\frac{\partial^2 FD_0}{\partial \xptwo^2}}
\providecommand{\ddFDdxponetwo}{\frac{\partial^2 FD_0}{\partial \xpone \partial \xptwo}}

\providecommand{\dHTdxpone}{\frac{\partial \HT}{\partial \xpone}}
\providecommand{\dHTdxptwo}{\frac{\partial \HT}{\partial \xptwo}}
\providecommand{\dHTHdx}{\frac{\partial (\HT H)}{\partial \X}}
\providecommand{\dHTHdxi}{\frac{\partial (\HT H)}{\partial \X_i}}
\providecommand{\dHTHdxone}{\frac{\partial (\HT H)}{\partial \X_1}}
\providecommand{\dHTHdxN}{\frac{\partial (\HT H)}{\partial \X_{N}}}
\providecommand{\dHTHdxNone}{\frac{\partial (\HT H)}{\partial \X_{N+1}}}
\providecommand{\dHTHdxtwoN}{\frac{\partial (\HT H)}{\partial \X_{2N}}}
\providecommand{\dHTdxi}{\frac{\partial \HT}{\partial \X_{i}}}
\providecommand{\dHdxi}{\frac{\partial H}{\partial \X_{i}}}
\providecommand{\dHTHdxtwoN}{\frac{\partial (\HT H)}{\partial \X_{2N}}}

\providecommand{\dHHTinvdx}{\frac{\partial (\HT H)^-1}{\partial \X}}
\providecommand{\dHdX}{\frac{\partial H}{\partial \X}}
\providecommand{\dHdxpone}{\frac{\partial H}{\partial \xpone}}
\providecommand{\dHdxptwo}{\frac{\partial H}{\partial \xptwo}}
\providecommand{\ddHddxpone}{\frac{\partial^2 H}{\partial \xpone^2}}
\providecommand{\dHdotdxpone}{\frac{\partial \Hdot}{\partial \xpone}}
\providecommand{\dHdotTdxpone}{\frac{\partial \Hdot^T}{\partial \xpone}}
\providecommand{\dCdotdxpone}{\frac{\partial \Cdot}{\partial \xpone}}
\providecommand{\dHdotdxdpone}{\frac{\partial \Hdot}{\partial \xdpone}}
\providecommand{\dHdotTdxdpone}{\frac{\partial \Hdot^T}{\partial \xdpone}}
\providecommand{\dCdotdxdpone}{\frac{\partial \Cdot}{\partial \xdpone}}
\providecommand{\dCdotdxdptwo}{\frac{\partial \Cdot}{\partial \xdptwo}}

\providecommand{\dbdt}{{\frac{\partial b}{\partial t}}}
\providecommand{\dbdx}{{\frac{\partial b}{\partial \X_s(t)}}}
\providecommand{\dbdxt}{{\frac{\partial b}{\partial \X_s(t)}}}
\providecommand{\dbdxg}{\frac{\partial b(g_j(\X_s(t),\Xd_s(t)))}{\partial \X_s(t)}}
\providecommand{\dbdxdg}{\frac{\partial b(g_j(\X_s(t),\Xd_s(t)))}{\partial \Xd_s(t)}}
\providecommand{\dbdxgu}{\frac{\partial b(h_i(\uu_s(t)))}{\partial \X_s(t)}}
\providecommand{\dbdxdgu}{\frac{\partial b(h_i(\uu_s(t)))}{\partial \Xd_s(t)}}
\providecommand{\dbdxtg}{{\frac{\partial b(g_j(\X_s(t)))}{\partial \X_s(t)}}}
\providecommand{\gx}{g_j(\X_s(t),\Xd_s(t))}
\providecommand{\gu}{h_i(\uu_s(t))}
\providecommand{\gsqx}{({g_j}^2(\X_s(t),\Xd_s(t)))}
\providecommand{\Sgx}{S(g_j(\X_s(t),\Xd_s(t)))}
\providecommand{\Sgu}{S(h_i(\uu_s(t)))}
\providecommand{\dgdxx}{\frac{\partial g_j(\X_s(t),\Xd_s(t))}{\partial \X_s(t)}}
\providecommand{\dgdxdx}{\frac{\partial g_j(\X_s(t),\Xd_s(t))}{\partial \Xd_s(t)}}

\providecommand{\dSdgx}{\frac{\partial S}{\partial g_j}(\X_s(t),\Xd_s(t))}
\providecommand{\dgdxdx}{\frac{\partial g_j(\X_s(t),\Xd_s(t))}{\partial \Xd_s}}

\providecommand{\ddtdgdxdx}{\frac{d}{dt} \bigg(\frac{\partial g_j(\X_s(t),\Xd_s)}{\partial \Xd_s} \bigg) }
\providecommand{\dgdtx}{\frac{d g_j(\X_s(t),\Xd_s)}{dt}}
\providecommand{\dSdtx}{\frac{d S}{dt}}
\providecommand{\ddtdSdgx}{\frac{d}{dt} \bigg( \frac{\partial S}{\partial g_j}(\X_s(t),\Xd_s) \bigg)}

\providecommand{\g}{h_i}
\providecommand{\gsq}{{h_i}^2}
\providecommand{\Sg}{S(h_i)}
\providecommand{\dgdx}{\frac{\partial h_i}{\partial \X_s}}
\providecommand{\dgdxd}{\frac{\partial h_i}{\partial \Xd_s}}

\providecommand{\dSdg}{\frac{\partial S}{\partial h_i}}
\providecommand{\dgdxd}{\frac{\partial h_i}{\partial \Xd_s}}

\providecommand{\ddtdgdxd}{\frac{d}{dt} \bigg(\frac{\partial h_i}{\partial \Xd_s} \bigg) }
\providecommand{\dgdt}{\frac{d h_i}{dt}}
\providecommand{\dSdt}{\frac{d S}{dt}}
\providecommand{\ddtdSdg}{\frac{d}{dt} \bigg( \frac{\partial S}{\partial h_i} \bigg)}

\providecommand{\dgdu}{{\frac{\partial h_i}{\partial u_s}}}
\providecommand{\dudx}{{\frac{\partial u_s}{\partial \X_s}}}
\providecommand{\dudxd}{{\frac{\partial u_s}{\partial \Xd_s}}}

\providecommand{\lfull}{c(\x(t), \xd(t), \uu(t))}
\providecommand{\lfullus}{c(\X_s(t), \Xd_s(t), \uu_s(t))}

\providecommand{\dCdxpone}{\frac{\partial C}{\partial \xpone}}
\providecommand{\dCdxptwo}{\frac{\partial C}{\partial \xptwo}}

\providecommand{\ddCddxpone}{\frac{\partial^2 C}{\partial \xpone^2}}
\providecommand{\ddCddxptwo}{\frac{\partial^2 C}{\partial \xptwo^2}}
\providecommand{\ddCdxponetwo}{\frac{\partial^2 C}{\partial \xpone \partial \xptwo}}
\providecommand{\ddCdxptwoone}{\frac{\partial^2 C}{\partial \xptwo \partial \xpone}}

\providecommand{\dCdotdxpone}{\frac{\partial \Cdot}{\partial \xpone}}
\providecommand{\dCdotdxptwo}{\frac{\partial \Cdot}{\partial \xptwo}}

\providecommand{\ddIDddxpone}{\frac{\partial^2 ID}{\partial \xpone^2}}
\providecommand{\ddIDddxptwo}{\frac{\partial^2 ID}{\partial \xptwo^2}}
\providecommand{\ddIDdxponetwo}{\frac{\partial^2 ID}{\partial \xpone \partial \xptwo}}

\providecommand{\FDuzero}{FD_{u=0}}

\providecommand{\dHTHdxpone}{\frac{\partial (\HT H)}{\partial \xpone}}
\providecommand{\dHTHdxptwo}{\frac{\partial (\HT H)}{\partial \xptwo}}
\providecommand{\dHTHinvdxpone}{\frac{\partial (\HT H)^{-1}}{\partial \xpone}}
\providecommand{\dHTHinvdxptwo}{\frac{\partial (\HT H)^{-1}}{\partial \xptwo}}
\providecommand{\dHinvdxpone}{\frac{\partial (H)^{-1}}{\partial \xpone}}
\providecommand{\dHinvdxptwo}{\frac{\partial (H)^{-1}}{\partial \xptwo}}
\providecommand{\HTH}{\HT H}
\providecommand{\HTHinv}{(\HT H)^{-1}}
\providecommand{\dgammadxpone}{\frac{\partial \gamma}{\partial \xpone}}
\providecommand{\dgammadxptwo}{\frac{\partial \gamma}{\partial \xptwo}}
\providecommand{\dgammadxdpone}{\frac{\partial \gamma}{\partial \xdpone}}
\providecommand{\dgammadxdptwo}{\frac{\partial \gamma}{\partial \xdptwo}}
\providecommand{\dgammadxddpone}{\frac{\partial \gamma}{\partial \xddpone}}
\providecommand{\dgammadxddptwo}{\frac{\partial \gamma}{\partial \xddptwo}}

\providecommand{\dalphaonedxpone}{\frac{\partial \alpha_1}{\partial \xpone}}
\providecommand{\dalphaonedxptwo}{\frac{\partial \alpha_1}{\partial \xptwo}}
\providecommand{\dalphatwodxpone}{\frac{\partial \alpha_2}{\partial \xpone}}
\providecommand{\dalphatwodxptwo}{\frac{\partial \alpha_2}{\partial \xptwo}}
\providecommand{\dalphaonedxdpone}{\frac{\partial \alpha_1}{\partial \xdpone}}
\providecommand{\dalphaonedxdptwo}{\frac{\partial \alpha_1}{\:\partial \xdptwo}}
\providecommand{\dalphatwodxdpone}{\frac{\partial \alpha_2}{\partial \xdpone}}
\providecommand{\dalphatwodxdptwo}{\frac{\partial \alpha_2}{\partial \xdptwo}}
\providecommand{\dGammadxpone}{\frac{\partial \Gamma}{\partial \xpone}}
\providecommand{\dGammadxptwo}{\frac{\partial \Gamma}{\partial \xptwo}}
\providecommand{\dGammadxdpone}{\frac{\partial \Gamma}{\partial \xdpone}}
\providecommand{\dGammadxdptwo}{\frac{\partial \Gamma}{\partial \xdptwo}}

\providecommand{\dOmegaonedx}{\frac{d \Omega_1}{d \X}}
\providecommand{\dOmegatwodx}{\frac{d \Omega_2}{d \X}}
\providecommand{\dOmegathreedx}{\frac{d \Omega_3}{d \X}}
\providecommand{\dOmegafourdx}{\frac{d \Omega_4}{d \X}}
\providecommand{\dOmegaonedxd}{\frac{d \Omega_1}{d \Xd}}
\providecommand{\dOmegatwodxd}{\frac{d \Omega_2}{d \Xd}}
\providecommand{\dOmegathreedxd}{\frac{d \Omega_3}{d \Xd}}
\providecommand{\dOmegafourdxd}{\frac{d \Omega_4}{d \Xd}}

\providecommand{\dxpone}{\frac{d}{d \xpone}}
\providecommand{\dxptwo}{\frac{d}{d \xptwo}}
\providecommand{\dxpone}{\frac{d}{d \xpone}}
\providecommand{\dxptwo}{\frac{d}{d \xptwo}}

\providecommand{\ccons}{c_{\text{cons}}}

\providecommand{\Omegak}{\Omega_{k_{i}}}
\providecommand{\Xcont}{\mathcal{X}_{cont}}
\providecommand{\Xinit}{\X_{init}}
\providecommand{\ubar}{\bar{\uu}}
\providecommand{\kd}{k_d}
\providecommand{\kl}{k_c}

\providecommand{\Uff}{u_{ff}}
\providecommand{\xint}{\tilde{\X}}
\providecommand{\xdint}{\dot{\tilde{\X}}}
\providecommand{\qint}{\q_{int}}
\providecommand{\qdint}{\qd_{int}}
\providecommand{\ubarff}{\bar{\uu}_{ff}}
\providecommand{\Ufb}{u_{fb}}



\setlength{\textfloatsep}{18pt}

\makeatletter
\let\@oldmaketitle\@maketitle
    \renewcommand{\@maketitle}{\@oldmaketitle
    \centering
    \includegraphics[trim={0cm, 0cm, 0cm, 0cm},clip,width=2.\columnwidth,angle=0]{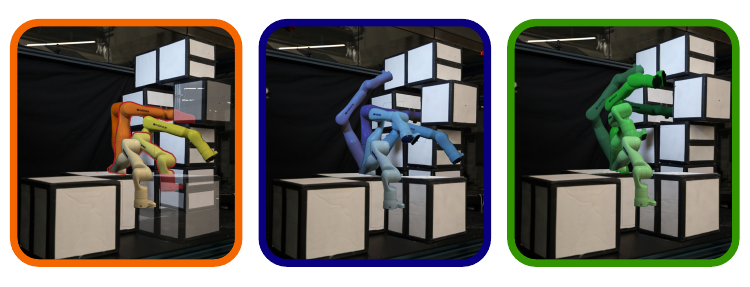}
    \captionof{figure}{
     This paper introduces \methodnamenew{}, a Phase 1 - Phase 2 Affine Geometric Heat Flow (AGHF) framework, to rapidly solve optimal control problems while respecting robot constraints and avoiding obstacles. 
    It begins with an initial trajectory (shown in \textcolor{init_traj_orange}{\textbf{orange}} with the color gradient illustrating the evolution in time starting from darkest and going to lightest) that may violate constraints (e.g., the second and fourth pose of the arm are in collision with the boxes and outlined in \textcolor{obs_red}{\textbf{red}}).
   If the initial trajectory is infeasible, \methodnamenew{} enters Phase 1, where it evolves the trajectory into a trajectory that satisfies all constraints (e.g., in the \textcolor{phase1_traj_blue}{\textbf{blue}} trajectory, the Kinova arm has been moved out of collision with the boxes).
    Once the trajectory satisfies all constraints, Phase 2 begins, optimizing the motion to minimize a user-specified cost function while maintaining feasibility (optimized trajectory shown \textcolor{final_traj_green}{\textbf{green}}).
    \methodnamenew{} optimizes the trajectory to reach a target configuration while avoiding the obstacles while considering the full dynamical model of the arm. 
    Note that optimal control (including Phase 1 and Phase 2) for this 14 dimensional state space model is completed within $3s$ while satisfying input, state, and collision avoidance constraints. 
    }
    \label{fig:AGHF-intro}
    \setcounter{figure}{1}
    \vspace*{-0.25cm}
  }
\makeatother

\maketitle
 \vspace*{-0.5cm} 
\thispagestyle{empty}
\pagestyle{plain}
\begin{abstract}
    The generation of optimal trajectories for high-dimensional robotic systems under constraints remains computationally challenging due to the need to simultaneously satisfy dynamic feasibility, input limits, and task-specific objectives while searching over high-dimensional spaces. 
    Recent approaches using the Affine Geometric Heat Flow (AGHF) Partial Differential Equation (PDE) have demonstrated promising results, generating dynamically feasible trajectories for complex systems like the Digit V3 humanoid within seconds. 
    These methods efficiently solve trajectory optimization problems over a two-dimensional domain by evolving an initial trajectory to minimize control effort. 
    However, these AGHF approaches are limited to a single type of optimal control problem (i.e., minimizing the integral of squared control norms) and typically require initial guesses that satisfy constraints to ensure satisfactory convergence.
    These limitations restrict the potential utility of the AGHF PDE especially when trying to synthesize trajectories for robotic systems.
    This paper generalizes the AGHF formulation to accommodate arbitrary cost functions, significantly expanding the classes of trajectories that can be generated. 
    This work also introduces a Phase 1 - Phase 2 Algorithm that enables the use of constraint-violating initial guesses while guaranteeing satisfactory convergence.
    The effectiveness of the proposed method is demonstrated through comparative evaluations against state-of-the-art techniques across various dynamical systems and challenging trajectory generation problems.
    Project Page: \href{https://roahmlab.github.io/BLAZE/}{https://roahmlab.github.io/BLAZE/}
\end{abstract}

\IEEEpeerreviewmaketitle

\section{Introduction}\label{sec:intro}

Optimal Control is a powerful tool for motion planning and control of advanced robotic systems.
For robust deployment in trajectory-based robotics-algorithms \cite{minicheetah, liuradius, kim_cheetah3, kuindersma2016optimization, michaux2024let, dragan2016}, an optimal control algorithm should be (1) computationally efficient to enable online planning, (2) capable of handling nonlinear dynamics and constraints inherent in real-world tasks, (3) scalable to high-dimensional platforms like humanoids and manipulators, and (4) reliably convergent to feasible solutions despite their initialization. 
However, balancing these criteria is challenging.
Current methods to address these challenges can be grouped into two primary approaches -- Dynamic Programming (DP) and Variational methods.
DP leverages Bellman's Principle of Optimality to derive value functions and optimal policies. 
In particular, solving the Hamilton-Jacobi-Bellman (HJB) Partial Differential Equation (PDE) offers global optimality guarantees, but necessitates discretizing the entire state space, making it computationally unfeasible for systems beyond 5-6 dimensions \cite{HJB_survey, bui2022optimizeddp}.
\renewcommand\thefootnote{}\footnotetext{$^*$ These authors contributed equally to this work}\addtocounter{footnote}{-1}
Differential Dynamic Programming (DDP) methods, such as Crocoddyl \cite{crocoddyl2020}, mitigate these issues by applying DP techniques locally along a given trajectory.
Through iterative forward-backward passes, DDP variants achieve more rapid convergence, providing a practical compromise between optimality and computational tractability for high-dimensional robotic systems.
Although DDP methods are more computationally efficient than global approaches, they can exhibit poor convergence when initialized far from a local optimum.

Variational methods derive necessary optimality conditions via calculus of variations, offering an alternative route to solving control problems. 
Within this framework, direct methods discretize a continuous optimal control problem into a nonlinear program (NLP).
For example, direct collocation methods \cite{C-FROST, Tropic2020} approximate trajectories using polynomial functions. 
Although effective at handling complex constraints, these methods can be computationally intensive for high-dimensional systems and sensitive to discretization choices and initial guesses.

Recently, another PDE-based method using the Affine Geometric Heat Flow (AGHF) PDE has been proposed \cite{AGHF_OG, adu2024bringheatrapidtrajectory}.
Notably these methods have been shown to be able to generate trajectories for high dimensional systems faster than existing methods (e.g., on the order of seconds for systems with more than a 40 dimensional state space model).
These methods pose the trajectory optimization problem as the solution to a PDE that evolves an initial trajectory that may not be dynamically feasible into a final trajectory that is dynamically feasible while minimizing control input magnitudes. 
In contrast to the HJB PDE, the AGHF solution is defined over a two-dimensional domain irrespective of the system's dimension. 
This enables the AGHF PDE to achieve significant computational speedups while preserving dynamic feasibility in motion planning and incorporating various kinds of constraints.
Though these AGHF methods show promise for rapidly generating trajectories for high-dimensional systems, currently they have been restricted to considering only one kind of cost function -- the integral of the squared norm of the control inputs along the trajectory.
Additionally, to find solutions rapidly these methods often require an initial guess that does not violate any constraint other than the dynamic feasibility constraint. 

To address these limitations, this paper proposes BLAZE, a method that builds a generalized AGHF formulation that accommodates arbitrary cost functions, enabling the generation of diverse trajectories that were previously unattainable. 
This method enables one to rapidly compute dynamically feasible trajectories for complex, highly dynamic tasks for high dimensional systems by
leveraging spatial vector algebra, a pseudospectral method and a Phase 1-Phase 2 algorithm for the solving the AGHF PDE (as illustrated in Figure \ref{fig:AGHF-intro}).
The contributions of this paper are three-fold:
First, this paper illustrates how to formulate the AGHF Action Functional to solve optimal control problems with arbitrary cost functions (Section \ref{subsec:designing_L}). 
Second, this paper describes how to implement a Phase 1-Phase 2 style method that enables the AGHF initial guess to violate the constraints while still generating feasible solutions rapidly (Section \ref{sec:ph1ph2 AGHF}).
Third, this paper describes how to enforce input constraints in the AGHF (Section \ref{sec: AGHF input constraints}).
The utility of these contributions is illustrated by comparing the performance of the proposed method to state of the art trajectory optimization techniques and a hardware demonstration of the proposed method on the Kinova Gen3 robot.

The remainder of the paper is arranged as follows:
Section \ref{sec: notation} presents the background and introduces the relevant notation for the paper. 
Section \ref{sec:AGHF} introduces the AGHF and discusses the underlying theory associated with the AGHF. 
Section \ref{sec: AGHF input constraints} details how to incorporate constraints into the AGHF to enable actions like obstacle avoidance.
Section \ref{sec:experiments} evaluates the proposed algorithm’s efficiency through simulation comparisons with several state-of-the-art methods and validates its performance on a Kinova Gen3 hardware platform.

\section{Preliminaries}
\label{sec: notation}
This section introduces the notation used throughout this manuscript.
This paper is focused on performing trajectory optimization for robot systems whose dynamics can be written as follows:
\begin{equation}
\label{eq: manipulator dyn}
    \Hq\qdd(t) + C(\q(t), \qd(t)) = B \uu(t),
\end{equation}
where $q(t) \in \R^{N}$ is the configuration of the robot at time $t$, $\uu(t) \in \R^{m}$ is the input applied to the robot at time $t$,
$\Hq$ is the mass matrix, $C(\q(t), \qd(t))$ is the grouped Coriolis and gravity term and $B$ is the actuation matrix.
For convenience, let $\x(t)$ correspond to the vector of $\q(t)$ and $\qd(t)$. 
To be consistent with the notation in the rest of the paper we refer to the first $N$ and last $N$ components of $\x(t)$ as $\x_{P1}(t)$ and $\x_{P2}(t)$, respectively. 
Using these definitions, we can represent the dynamics of the robot \eqref{eq: manipulator dyn} as a control affine system:
\begin{equation}
\label{eq: control affine dyn}
    \xd(t) = \Fdx + \Fx \uu(t),
\end{equation}
where 
\begin{align}
\Fdx &= \begin{bmatrix}
      \label{eqn: Fd}
       \x_{P2}(t) \\
        -\Hinvq\Cq 
    \end{bmatrix} \\
\Fx &= \begin{bmatrix}
        \label{eqn: F}
        0_{N \times m} \\
        \Hinvq B 
    \end{bmatrix}
\end{align}
For convenience, we assume without loss of generality that we are interested in the evolution of the system for $t \in [0,T]$.
Lastly, let $L^2$ refer to the set of square integrable functions defined for $t \in [0,T]$.
Throughout this paper, we also utilize the following definition that is used throughout the calculus of variations because it allows one to describe how functions behave as they go to infinity:
\begin{defn}[Coercive Functions {\cite[Section 8.2]{evans2022partial}}]
A function $f:\mathbb{R}^n \to \mathbb{R}$ is called \emph{coercive} if there exist constants $\alpha > 0, \beta \geq 0$ such that $f(x) \geq \alpha \|x\|_{\infty} - \beta$ for all $x$.
\end{defn}

The objective of this paper is to develop an algorithm to construct a trajectory beginning from some user-specified initial condition, $\x_0$, and ending in some user-specified terminal condition, $\x_f$, while satisfying state and input constraints and the dynamics in \eqref{eq: control affine dyn} for all $t \in [0,T]$ all while minimizing an arbitrary cost function.
For convenience, let the zero superlevel set of functions $g_j$ for each $j \in \J$ and $h_i$ for each $i \in \I$ represent a collection of state and input inequality constraints where $\J \subset \N$ and $\I \subset \N$ are each finite sets.
Using these definitions, one can formulate the trajectory optimization problem:
\begin{equation*}
\begin{aligned}
& \underset{\uu \in L^2}{\text{inf}}
    & & \int_{0}^{T} \lfull  ~dt &&\quad \text{(OCP)} \\
& \text{s.t.}
& & \xd(t) = \Fdx + \Fx \uu(t),  &&\forall t \in [0, T], \\
& & & g_j(\x(t),\xd(t)) \leq 0 &&\forall t \in [0, T], \forall j \in \J, \\ 
& & & h_i(\uu(t)) \leq 0 &&\forall t \in [0, T], \forall i \in \I, \\
& & & \x(0) = \x_0,&& \\
& & & \x(T) = \x_f, &&
\end{aligned}
\end{equation*}
The goal of this paper is to develop a method to rapidly generate trajectories for high-dimensional systems while incorporating multiple constraints using the Affine Geometric Heat Flow Partial Differential Equation to solve (OCP).
To ensure the convergence of the AGHF, we make the following assumption:
\begin{assum}[Continuity of Dynamics]
\label{assum:dynamics continuity}
Both $F_d$ and $F$ are $C^2$, globally Lipschitz continuous (with constants $Y_1$ and $Y_2$ respectively), and $F$ has constant rank almost everywhere in $\R^n$.
Suppose $c$, $g_j$, and $h_i$ are $C^2$ in all of their arguments for all $j \in \J$ and $i \in \I$.
Additionally, assume that there exists a trajectory that satisfies the constraints of (OCP).
\end{assum}
\noindent Note that the dynamics of rigid-body robotic systems are smooth and Lipschitz continuous when restricted to a compact domain.
In addition, we do not assume that the trajectory that satisfies the constraints of (OCP) is given to the user. 
Rather we only assume its existence.

\section{The Affine Geometric Heat Flow Partial Differential Equation}
\label{sec:AGHF}
The Affine Geometric Heat Flow (AGHF) is a parabolic PDE framework for trajectory optimization that deforms an arbitrary initial trajectory (including one that does not satisfy the dynamics) into a dynamically feasible one while minimizing the squared control input of the trajectory.
In this section, we review the foundational AGHF theory and introduce novel theorems with proofs that generalize the AGHF framework to optimal control problems with arbitrary cost functions as posed in (OCP).
A more detailed treatment of the background knowledge on the AGHF can be found in \cite{AGHF_OG,liu2019homotopy,liu2020geometric, adu2024bringheatrapidtrajectory}.
\emph{Throughout this section, we assume that there are no inequality constraints in (OCP). 
The inequality constraint case is considered in Section \ref{sec: AGHF input constraints}.}

\subsection{Homotopies and the Action Functional}
\label{subsec: homotopy}
To describe the evolution of trajectories by the AGHF PDE, we begin by defining a homotopy: $\X:[0,T] \times [0,s_{max}] \to \R^{2N}$, that is twice differentiable with respect to its first argument and differentiable with respect to its second argument. 
For convenience, we denote $\X(t,s)$ by $\X_s(t)$ and we denote $\frac{\partial \X}{\partial t}(t,s)$ by $\dot{\X}(t,s)$ or $\dot{\X}_s(t)$.
In practice this homotopy can be any twice differentiable initial guess for a trajectory that starts at $\x_0$ and terminates at $\x_f$.
As is done in Section \ref{sec: notation}, we refer to the first $N$ and last $N$ components of $\X_s$ as $\X_{P1}$ and $\X_{P2}$, respectively. 
Additionally, let the first and second time derivatives of these states be defined as $\xdpone$, $\xdptwo$ and $\xddpone$, $\xddptwo$, respectively.

Next, define the \emph{Lagrangian} $L: \R^{2N} \times \R^{2N} \to \R$, which we assume is $C^2$ and twice differentiable with respect to any of its arguments.
Subsection \ref{subsec:designing_L} describes how to select $L$ to ensure that the AGHF minimizes the cost function in (OCP).
Finally, define the \emph{Action Functional}:
\begin{equation}
\label{eqn:action functional}
    \Act(\X_s) = \int_0^T  L(\X_s(t),\dot{\X}_s(t)) dt.
\end{equation}

Using these definitions, we can define the AGHF PDE:
\begin{defn}[Affine Geometric Heat Flow]
\label{defn:AGHF}
The \emph{Affine Geometric Heat Flow} is a parabolic partial differential equation defined as:
\begin{align}
\label{eqn:AGHF}
    \dxds(t,s) &= \Minv (\X(t,s)) \bigg( 
    \ddt \dldxds(\X_s(t), \dot{\X}_s(t)) \nonumber \\
    &\quad - \dldxs(\X_s(t), \dot{\X}_s(t)) \bigg),
\end{align}
with the following boundary conditions:

\begin{equation}
\label{eqn: bcs aghf}
\begin{aligned}
    \X_s(0) &= \x_0,  \quad \forall s \in [0, s_{max}], \\ 
    \X_s(T) &= \x_f, \quad \forall s \in [0, s_{max}].
\end{aligned}
\end{equation}
where $M: \R^{2N} \to \mathbb{S}_{+}^{2N}$ is a user specified matrix valued function and $\mathbb{S}_{+}^{2N}$ is the set of positive semi-definite matrices.
\end{defn}
\noindent Note that this AGHF formulation differs from standard approaches in that here we allow $M(\X(t, s))$ to be an arbitrary positive semi-definite and invertible matrix.
This enables us to consider general cost functions.
The traditional AGHF formulation designs $M$ specifically to penalize evolution toward dynamically infeasible states while prioritizing control effort reduction.
However, as we show in this paper, for general cost functions, $M$, should be defined differently.

When solving the AGHF PDE, one begins by specifying an initial curve $\X_{init}:[0,T] \to \R^{2N}$.
As the AGHF PDE evolves forward in $s$, one can prove that the action functional is minimized. 
In addition, if during that evolution the AGHF converges to a curve where the right-hand side of the AGHF PDE is equal to $0$, then one has found a curve that extremizes the Action Functional.
Such a curve is called a \emph{steady state solution}.
We formalize these observations in the following Lemma that describes the convergence properties of the AGHF and whose proof can be found in Appendix \ref{sec: appendix aghf dissipation}.
\begin{lem}[Action Functional Along the AGHF Homotopy]
\label{lem:AGHF convergence}
Let $\X_s$ satisfy the AGHF PDE \eqref{eqn:AGHF}. 
Then, $\frac{d \Act(\X_s)}{ds} \leq 0$ for all $s$.
In addition, if the right hand side of the AGHF PDE when evaluated at $\X_{s^*}$ is equal to $0$ for some $s^* \in [0,s_{max})$, then $\frac{d\Act(\X_{s^*})}{ds} = 0$. 
\end{lem}
\noindent In other words, given the Action Functional \eqref{eqn:action functional}, as the AGHF PDE evolves forward in $s$, the Action Functional is minimized.
In addition, if during that evolution the AGHF converges to a curve where the right hand side of the AGHF PDE is equal to $0$, then one has found a \emph{steady state solution} to the AGHF.

\subsection{Solving the AGHF Rapidly}
\label{subsec:phlame_implementation}

Unlike the Hamilton-Jacobi-Bellman (HJB) PDE, which suffers from the curse of dimensionality due to its exponential scaling with state dimension, the AGHF PDE scales polynomially \cite{fan2019midair}.
This favorable scaling arises because the solution to the AGHF PDE has a two-dimensional domain and and has a range whose dimension grows linearly with the state dimension.
Because the AGHF solution has a two-dimensional domain, the AGHF PDE can be solved using the Method of Lines (MOL) \cite{schiesser2012numerical}.

The MOL discretizes the AGHF solution domain along one dimension (typically $t$) into a set of nodes, reformulating the PDE at each node as a system of coupled Ordinary Differential Equations (ODEs).
These coupled ODEs can then be solved in $s$ using numerical ODE solvers.
During each ODE solver step, the right-hand side (RHS) of the AGHF PDE must be evaluated at every node.
For high-dimensional systems, this process becomes computationally expensive, as it requires repeatedly computing the system dynamics and their derivatives.

In classical MOL AGHF implementations \cite{AGHF_OG}, the nodes are uniformly spaced, and the solution accuracy improves as more nodes are introduced. 
However, because the number of function evaluations scales linearly with the number of nodes, a finer discretization requires frequent evaluations of the AGHF RHS, significantly increasing computational cost.
Recent approaches \cite{adu2024bringheatrapidtrajectory} address this by employing a pseudospectral MOL that strategically reduces the number of required nodes while maintaining accuracy.
This pseudospectral formulation enables precise computation of time derivatives.
Additionally, one can accelerate the evaluation of the AGHF's RHS by utilizing spatial vector algebra and rigid body dynamics algorithms \cite{adu2024bringheatrapidtrajectory}.

\subsection{Ensuring $\Act$ Coincides with the (OCP) Cost Function}
\label{subsec:designing_L}

To ensure that the Action Functional, $\Act$, being minimized coincides with minimizing an arbitrary cost function, $\lfull$, of the (OCP) we must design $L$ carefully.
Before defining this action functional we first introduce one additional definition:
\begin{defn}[Control Extraction]
\label{defn:control_extraction}
Suppose $\Xs$ corresponds to a continuously differentiable trajectory at any $s \in [0, s_{max}]$. 
 Let $\uu_s:[0,T] \to \R^{m}$ be the \emph{extracted control input} given by:
    \begin{equation}
    \label{eqn:control extraction}
        \uu_s(t) = \begin{bmatrix} 0_{N \times  N} & I_{N \times N} \end{bmatrix}
        \bar{F}(\X_s(t))^{-1} (\dot{\X}_s(t) - \Fd(\X_s(t))).
    \end{equation}
    where
    \begin{equation}
    \label{eqn:fbar}
        \bar{F}(\X_s(t)) =  \begin{bmatrix}
            F_c(\X_s(t)) & F(\X_s(t))
            \end{bmatrix} \in \R^{2N \times 2N},
    \end{equation}
and $F_c(\X_s(t)) \in \R^{2N \times (2N-m)}$ is a differentiable matrix such that $\bar{F}$ is invertible for all $\X_s(t)$ in $\R^{2N}$.
\end{defn}
\noindent This definition describes how to synthesize a control input given a specific trajectory. 
For notational convenience, we have omitted the explicit dependency of $\uu_s$ on $\X_s$ and $\Xd_s$.
Note that $F_c$ in the above definition can be obtained using the Gram-Schmidt procedure.
We give an explicit example of how to construct $F_c$ when describing our experimental results in Section \ref{sec:experiments}.
Before moving on, we call attention to one important consideration: it remains unclear whether applying this extracted control input to the dynamical system \eqref{eq: control affine dyn} would reproduce the original trajectory $\X_s$ used in its creation.
This question is answered in Theorem \ref{thm:conv_dyn_feasible}. 

Given Definition \ref{defn:control_extraction}, we can define a Lagrangian and associated Action Functional that encode solutions to (OCP) as we describe next:
\begin{defn}[Lagrangian and Action Functional for (OCP)]
    \label{defn:act_func_ocp}
    Let $\uu_s:[0,T] \to \R^{m}$ be the extracted control input at some $s \in [0, s_{max}]$ using Definition \ref{defn:control_extraction}.
Define $L$ in terms of the extracted control input $u_s(t)$ as:
\begin{align}
\label{eqn:L(x)}
    \begin{split}
        L(\X_s(t), \Xd_s(t)) = \kd  \|\xdponest - \xptwost\|_2^2 \\
        & \hspace{-4.3cm} + \lfullus 
    \end{split}
\end{align}
where $\kd > 0$.
Then the corresponding Action Functional is given by
\begin{align}
\begin{split}
    \label{eqn:action functional k}
        \Act(\X_s) = \int_0^T \bigg( \kd  \|\xdponest - \xptwost\|_2^2 \\
            & \hspace{-4.3cm} +  \lfullus \bigg) dt 
\end{split}
\end{align}  
\end{defn}

\noindent In summary, for each $s$, the Action Functional with $L$ as described by Definition \ref{defn:act_func_ocp} consists of the integral of an arbitrary objective function, $\lfull$, plus the error between the velocity states, $\xptwo$, and the derivative of the position state, $\xdpone$.

For trajectories that satisfy the system dynamics \eqref{eq: control affine dyn}, this error term vanishes, reducing the Action Functional to the integral of $\lfull$.
Notably, as Lemma \ref{lem:AGHF convergence} shows, as the AGHF evolves the Action Functional decreases, which coincides with the objective function in (OCP) for dynamically feasible trajectories.

However, it is unclear whether the solution generated by the AGHF PDE eventually satisfies the system dynamics \eqref{eq: control affine dyn}.
The next result whose proof can be found in Appendix \ref{sec: appendix aghf convergence} resolves this concern:
\begin{thm}[AGHF Generates Feasible Trajectories]
\label{thm:conv_dyn_feasible}
Consider a system governed by the dynamics in \eqref{eq: control affine dyn}, with an initial state $\x_0$ and desired final state $\x_f$.
Suppose Assumption \ref{assum:dynamics continuity} is satisfied, $F_c$ in Definition \ref{defn:control_extraction} is a continuous function, and the cost function in the Action Functional \eqref{eqn:action functional k} is coercive.
Then there exists $C_1, C_2,$ and $C_3 > 0$, such that for any $k_d > 0$, there exists an open set $\Omega_{k_d} \subset \{ x \in C^1([0,T] \to \mathbb{R}^{2N}) \mid x(0) = \x_0, x(T) = \x_f \}$ such that as long as the initial curve to the AGHF PDE satisfies $x_{init} \in \Omega_{k_d}$ then for sufficiently large $s_{max}$ the integrated path $\tilde{x}$ using the extracted control input with $x_{s_{max}}$ plugged into the dynamics \eqref{eq: control affine dyn} satisfies the following error bound for any $t \in [0,T]$:
\begin{equation}
\label{eqn:error bound}
    \| \tilde{x}(t) - x_{s_{max}}(t)\|_{2} \leq \sqrt{ \frac{3 T C_1 C_2^2}{k_d} \exp{\bigg( 3 T (Y_1^2 T + Y_2^2 C_3) \bigg)} }
\end{equation}
\end{thm}

\noindent Theorem \ref{thm:conv_dyn_feasible} proves that for sufficiently large $\kd$ and $s_{max}$, the control input extracted from the solution to the AGHF PDE can be used to generate a trajectory that is arbitrarily close to a dynamically feasible trajectory of \eqref{eq: control affine dyn}   and provides an explicit bound on how close the trajectory obtained by applying \eqref{eqn:control extraction} is to a feasible trajectory of (OCP).
With this result, we have an AGHF formulation that tries to minimize the objective function in (OCP) and enables one to generate dynamically feasible trajectories.

\begin{rem}[Relationship to Previous Lagrangians]
    Note that in Definition \ref{defn:act_func_ocp} if $L$ is replaced by
    \begin{align}
        \label{eqn:lagrangian}
        L(\X_s(t), \dot{\X}_s(t)) = 
        &\ \big(\dot{\X}_s(t) - F_d(\X_s(t))\big)^T G(\X_s(t)) \nonumber \\
        &\ \cdot \big(\dot{\X}_s(t) - F_d(\X_s(t))\big),
    \end{align}
where $G$ is given by
\begin{equation}
     G = \begin{bmatrix}
        kI_{N \times N} & 0_{N \times N} \\
        0_{N \times N} & H^T H
        \end{bmatrix} \in \R^{2N \times 2N}
\end{equation}
and $M=G$, then the resulting Action Functional and AGHF correspond to the standard formulation used in the literature \cite{AGHF_OG,adu2024bringheatrapidtrajectory}, which minimizes the squared control input (i.e $\lfullus$ is $\|\uu_s(t) \|_2^2$).
However, the formulation in Definition \ref{defn:act_func_ocp} along with Theorem \ref{thm:conv_dyn_feasible} generalizes this framework to accommodate a broader class of optimal control problems.
\end{rem}

\section{Incorporating Arbitrary Constraints into the AGHF}
\label{sec: AGHF input constraints}

The previous section assumed that there were no constraints in (OCP) other than the dynamics constraints.
This section discusses how to enforce general state and input constraints.

\subsection{Constrained Lagrangian}
\label{subsec: AGHF constraint}

We incorporate constraints into the AGHF by using a penalty term in the Lagrangian:
\begin{defn}[Constrained Lagrangian]
\label{defn: constraint lagrangian}
Let $k_{cons}$ be some large positive real number, and let $g_j(\X(t), \Xd(t))$ be the $j$-th inequality constraint evaluated at $\X(t)$ and $\Xd(t)$.
Let the \emph{Constrained Lagrangian} denoted by $L_{cons}$ be defined as:
\begin{equation}
\label{eq: constraint lagrangian}
    L_{cons}(\X(t),\Xd(t))  = L(\X(t),\Xd(t))  + \sum_{j \in \J} b(g_j(\X(t),\Xd(t)))
\end{equation}
where
\begin{equation}
\label{eq: constraint penalty}
    b(g_j(\X(t),\Xd(t))) = k_{cons} \cdot (g_j(\X(t),\Xd(t)))^2 \cdot S(g_j(\X(t),\Xd(t))),
\end{equation}
where $S:\R \to \R$ is defined as follows:
\begin{enumerate}
    \item $S:\R \to \R$ is a positive, differentiable function, 
    \item $S(g_j(\X(t),\Xd(t))) = 0$ when $g_j(\X(t),\Xd(t)) \leq 0$ and
    \item $S(g_j(\X(t),\Xd(t))) = 1$ when $g_j(\X(t),\Xd(t)) > 0$.
\end{enumerate}
\end{defn}
\noindent The function $S$ ensures that the penalty term is activated only when the AGHF trajectory moves towards constraint violation.
An example of such a function is described during the description of the experiments in Section \ref{sec:experiments}.
With a sufficiently large $k_{cons}$, the penalty term steers the trajectory away from infeasible regions, preventing convergence to solutions that violate constraints. 

Using the Constrained Lagrangian, one can construct a corresponding AGHF by applying Definition \ref{defn:AGHF}:
\begin{lem}[AGHF for a Constrained Lagrangian]
\label{lem:constrained_AGHF}
Consider a Constrained Lagrangian, $L_{cons}$, as in Definition \ref{defn: constraint lagrangian}. 
Then the AGHF PDE is given by:
\label{eqn: constraint aghf}
 \begin{align}
 \begin{split}
     \dxds = \Minv(\X_s(t)) \bigg(\ddt \dldxds(\X_s(t),\Xd_s(t))  - \dldxs(\X_s(t),\Xd_s(t)) \\ 
    & \hspace{-9cm} +\sum_{j \in \J} \bigg(\ddt \dbdxdg - \dbdxg \bigg) \bigg )
 \end{split}
 \end{align}
\end{lem}
\noindent Notice that $L_{cons}$, still satisfies the properties of the Lagrangian introduced in Section \ref{subsec: homotopy}.
As a result, the convergence properties shown in Lemma \ref{lem:AGHF convergence} apply to the Constrained Lagrangian.
If the function $b$ in the definition of $L_{cons}$ is coercive, then the results of Theorem \ref{thm:conv_dyn_feasible} also apply to ensure that a dynamically feasible trajectory is found. 
Note this does not guarantee that the inequality constraints within (OCP) are  eventually satisfied. 
This is particularly an issue when the initial trajectory passed to the homotopy violates the constraints. 
We address this particular challenge in Section \ref{sec:ph1ph2 AGHF}.

\subsection{Incorporating Input Constraints in the Lagrangian}
\label{subsec: input constraints analytical}

This section discusses how to rapidly compute the derivatives of the input penalties which are required to enforce input constraints.
Before introducing this formulation, we briefly describe how alternative AGHF formulations have enforced input constraints. 
Existing approaches to enforce input constraints using the AGHF require augmenting the state by treating inputs as additional states, and enforcing the input constraints as state constraints\cite{AGHF_OG}. 
For high-dimensional robotic systems, this approach substantially increases the dimension of state space of the system and, consequently, the dimension of the AGHF PDE. 

To avoid this issue, we instead enforce the input constraints by appending $L_{cons}$ with the term $\sum_{i\in \I} b(h_i(u_s(t))$.
However to construct the AGHF as in Lemma \ref{lem:constrained_AGHF} for this even larger Constrained Lagrangian, one must be able to compute partial derivatives of $b(h_i(u_s(t))$ with respect to $\X_s(t)$ and $\dot{\X}_s(t)$. 
This requires then applying the chain rule and computing partial derivatives of $u_s(t)$ with respect to $\X_s(t)$ and $\dot{\X}_s(t)$. 
This can be challenging for high-dimensional robotic systems. 
Fortunately, one can leverage the following lemma, whose proof can be found in Appendix \ref{sec: appendix control RNEA}:

 \begin{lem}
    \label{lem: control inverse dyn}
     Suppose dynamics are given by \eqref{eq: manipulator dyn} where $B=I$ then the control extraction formula \eqref{eqn:control extraction} is equivalent to inverse dynamics and $u_s$ can be computed directly using inverse dynamics.
 \end{lem}
\noindent 
 From Lemma \ref{lem: control inverse dyn} it follows that one can compute $u_s$ using Rigid Body Dynamics Algorithms, such as the Recursive Newton-Euler Algorithm (RNEA).
 This can be used to efficiently compute $u_s(t)$ using $\X_s(t)$ and $\Xd_s(t))$ \cite{featherstone2014rigid}.
 By employing an extended version of RNEA that recursively applies the chain rule \cite{carpentierderivs} we can also rapidly evaluate the derivatives of the control inputs $\dudx$ and $\dudxd$.
\section{Designing a Phase 1 -Phase 2 Algorithm using the AGHF}
\label{sec:ph1ph2 AGHF}

Similar to other trajectory optimization methods, AGHF requires $\Xinit$ as an initial guess.
As detailed in Section \ref{sec:AGHF}, the AGHF PDE transforms this initial trajectory $\Xinit$ into an optimal final trajectory.  
If $\Xinit$ does not satisfy the optimal control problem's constraints, the AGHF solver must address constraint violations, leading to larger penalties that make the AGHF evaluate to larger and larger values, which can cause the ODE solver to take smaller steps to ensure solution accuracy, which increases the convergence time.
Previous approaches \cite{adu2024bringheatrapidtrajectory} mitigate this issue by providing initial guesses that satisfy the constraints, facilitating faster AGHF convergence. 
However, designing such feasible initial trajectories is challenging for systems with many complex constraints such as input constraints.

To enhance the efficiency of the AGHF and enable rapid convergence, we propose a Phase 1- Phase 2 Algorithm using the AGHF. 
Phase 1-Phase 2 Methods are used in optimization to first drive an initial guess into a feasible region (Phase 1) before optimizing the objective function while maintaining feasibility (Phase 2) \cite{polak2012optimization}. 
This approach aids an optimization process in beginning from a well-conditioned, constraint-satisfying state.
This results in improved convergence properties and computational efficiency.
To implement this within AGHF, we formulate two distinct Action Functionals -- one for each phase. 

In Phase 1, the goal is to guide the initial trajectory into the feasible set while promoting dynamic feasibility.
Leveraging Definition \ref{defn:act_func_ocp}, we construct an Action functional that primarily penalizes constraint violations to achieve this, with an additional term to encourage dynamic feasibility.
In Phase 2, we apply the generalized AGHF Action Functional with constraints \eqref{eqn: constraint aghf} to maintain feasibility while minimizing the objective cost.

We first introduce the form of the Phase 1 Action Functional:
\begin{defn}\label{defn:AGHF_Phase1}
Let the \emph{Phase 1 Action Functional} for a general state constraint $\gx$ be defined by
\begin{multline}
\label{eqn:action functional ph1}
    \Act(\X_s) = \int_0^T \kd \|\xdpone(s,t) - \xptwo(s,t)\|_2^2 \\ + \sum_{j\in \J}b(g_j(\X_s(t),\Xd_s(t)))
\end{multline}
and let the corresponding AGHF RHS be expressed as:
\begin{align}
\label{eqn:AGHF_ph1}
\begin{split}
    \dxds(t,s) = I_{2N \times 2N} \cdot\\
    & \hspace{-2cm} \bigg( 
    \ddt \dlddxds(\X_s(t),\Xd_s(t)) - \dlddxs(\X_s(t),\Xd_s(t)) \bigg) \bigg) \\
    & \hspace{-3cm} +\sum_{j \in \J} \bigg(\ddt \dbdxdg - \dbdxg \bigg),
\end{split}
\end{align}
where $I_{2N \times 2N}$ is the identity matrix and 
\begin{align}
    \begin{split}
        L_d(\X_s(t),\Xd_s(t)) = \big(\dot{\X}_s(t) - F_d(\X_s(t))^T \cdot \\
        & \hspace{-4cm} \begin{bmatrix}
        \kd I_{N \times N} & 0_{N \times N} \\
        0_{N \times N} & 0_{N \times N}
        \end{bmatrix} \big(\dot{\X}_s(t) - F_d(\X_s(t)).
    \end{split}
\end{align}
\end{defn}
Note that one can extend this definition to incorporate input constraints $\gu$.
Phase 2 utilizes the AGHF RHS from\eqref{eqn: constraint aghf}, initializing the trajectory evolution with the solution obtained from Phase1.
Given a sufficiently large penalty parameter $k_{cons}$ and evolution time $s_{max}$, where $k_{cons} > \kd$, the Phase 1 trajectory evolves towards a trajectory that minimizes constraint violation while promoting some dynamic feasibility.

Phase 2 is initialized with the final trajectory from Phase 1 and proceeds by optimizing the objective function while maintaining feasibility. 
As shown in Section \ref{subsec: AGHF constraint}, the constraint penalty term ensures that for sufficiently large $k_{cons}$ and $s_{max}$, where $k_{cons} > \kd$ if the trajectory starts in the feasible set, it will remain within the feasible set throughout the evolution.
Consequently, as the AGHF PDE progresses, the trajectory is driven toward a minimizer of the specified cost function that is dynamically feasible while satisfying all constraints.
\section{Experiments and Results}
\label{sec:experiments}

This section evaluates the speed and performance of \methodnamenew{} on a variety of scenarios and robot platforms.
We begin by explaining the experimental setup. 
Next, we evaluate the scalability of \methodnamenew{} when compared to the state of the art trajectory optimization algorithm. 
Then, we investigate the performance of the methods in scenarios where obstacles are present and then describe the translation of these results onto a real-world platform.
We conclude the section by summarizing the different evaluations.

\subsection{Implementation of \methodnamenew{}}

Our numerical implementation of \methodnamenew{} follows the pseudospectral MOL approach adopted in \cite{adu2024bringheatrapidtrajectory} which is summarized in Section \ref{subsec:phlame_implementation}.
 Throughout these experiments, for \methodnamenew{} we choose the following activation function $S(g_j(\X, \Xd))$ for state constraints that satisfies the properties highlighted in Definition \ref{defn: constraint lagrangian}:
\begin{equation}
\label{eq: barrier lagrangian}
    S(g_j(\X, \Xd)) = \frac{1}{2} + \frac{1}{2} \tanh(\ccons \cdot g_j(\X, \Xd))
\end{equation}
where $\ccons$ is a hyper-parameter that determines how fast $S(g_j(\X))$ transitions from 0 to 1, once the constraint is violated.
Note that we apply the same activation function for the input constraints as well, where in that case the activation is given by $S(g_j(\uu))$ instead. 
Note that the Constrained Lagrangian we choose in Phase 1 of our implementation requires that we satisfy input box-constraints for each of the examples described below.
Whereas in Phase 2, we try minimize the square of the control effort.
In either instance, note that the Lagrangian satisfies the coercive requirement due to the definition of the extracted control input (Definition \ref{defn:control_extraction}).

\subsection{Experimental Setup}
  We evaluate \methodnamenew{} and compare it to the DDP methods Crocoddyl \cite{crocoddyl2020} and Aligator \cite{aligator}, and RAPTOR \cite{zhang2024rapidrobusttrajectoryoptimization} which uses IPOPT as its backend optimizer. 
  Note we compare against these methods because they each are able to perform optimal control while considering the full order dynamics of a robot.
  We solve a number of fixed time and fixed initial and final state optimization problems.
 In all the problems, we enforce input constraints and state constraints and in problems with obstacles we also enforce obstacle avoidance constraints.
 All the experiments were run on an Ubuntu 22.04 machine with an AMD EPYC 7742 64-Core @ 256x 2.25GHz CPU.

\subsubsection{Selection of Parameters} The solution produced by any of the mentioned trajectory optimization algorithms depends heavily on the selection of initial guess and on the parameters of the solver. 
This section explains the rationale used to choose the parameters.

For most experiments we use the parameters reported in the grid search results of \cite{adu2024bringheatrapidtrajectory} with the following caveat.
When the experiments are qualitatively different from the ones considered in that paper (e.g., novel constraints or types of obstacles), we sequentially vary the parameters until finding parameters those that yield a feasible solution.
If the previous procedure fails, we run a small grid search around the parameters reported in \cite{adu2024bringheatrapidtrajectory}.
Appendix \ref{appendix: experiments parameters} summarizes all the parameters chosen for the different methods for all the experiments and explain the meaning of each.

\subsubsection{Evaluating Success or Failure}
\label{subsec: success}
Each tested numerical method generates an open-loop control input, $\uu^*:[0,T] \to \R^m$ and corresponding system trajectory, $\X^*:[0,T] \to \R^{2N}$.
 However, whether the robot can accurately track these computed solutions in practice is inobvious.
 Thus, to fairly evaluate constraint satisfaction and account for integration errors when applying the open-loop control, we integrate forward using the following feedback controller:
 \begin{equation}
    \Ufb(t) = \uu^*(t) +  k_p (\X^*_{P1}(t) - \q(t)) + k_v (\X^*_{P2}(t) - \qd(t)),
\label{eq: feed forward control}
\end{equation}
where $k_p = k_v = 100$ for each experiment, and $\q(t)$ and $\qd(t)$ represent the robot’s position and velocity at time $t$, respectively. 
The system dynamics are then integrated forward using \eqref{eq: feed forward control} to assess tracking accuracy and constraint adherence.

In obstacle-free experiments, a solution is considered \emph{successful} if the infinity norm of the error between the forward-integrated final state and $\X_f$ is below a fixed threshold $\epsilon = 0.05$ and the level of constraint violation is of the forward integrated solution is less than 5\% of the maximum constraint bound (e.g. for $u_{max} = 5$, and $u_{min} = -5$ then the control must satisfy $-5.25 \leq u(t) \leq 5.25,~ \forall t \in [0, T]$).
In experiments with obstacles, a trial is considered \emph{successful} if it meets the previous criteria and, additionally, the joint positions of the forward integrated solution, sampled at a time resolution of $\Delta t = 10^{-2}$s, remain outside of the obstacles at each joint frame.

\subsubsection{Aligning Initial Guesses}
\label{subsubsection: initial guess}
For all the experiments, each method is given the same initial guess $\Xinit$ to ensure a fair comparison across methods. 
Because Crocoddyl and Aligator, both DDP-based methods, optimize over control inputs as decision variables, we provide them with an initial guess of $\uu_{init}:[0,T] \to \R^N$.
This is obtained via inverse dynamics using position $q:[0,T] \to \R^N$, velocity $\qd:[0,T] \to \R^N$ and acceleration $\qdd:[0,T] \to \R^N$ extracted from $\Xinit$.
By default $\Xinit$ provides $q$ and $\qd$. 
To compute the acceleration trajectory $\qdd:[0,T] \to \R^N$, we fit a Chebyshev polynomial to $\Xinit$ and apply the Chebyshev differentiation matrix $D:\mathbb{R}^{p+1} \to \mathbb{R}^{p+1}$ as described in \cite[(21.2)]{trefethen2019approximation}.
Specifically, the differentiation matrix $D$ approximates the derivative of the fitted polynomial, allowing the computation of $\qdd$ from the polynomial coefficients.
The resulting trajectories $\q$, $\qd$ and $\qdd$ are then utilized in the inverse dynamics computation to generate $\uu_{init}$.
In contrast, RAPTOR formulates trajectory optimization with Bezier polynomial coefficients as decision variables. To ensure a consistent initial guess, we fit a Bezier polynomial to $\Xinit$.
$\Xinit$ and use the fitted polynomial's coefficients as RAPTOR’s initial input.

\subsection{BLAZE Scalability with Increasing System Dimension}

This section compares \methodnamenew{}, Crocoddyl, Aligator and RAPTOR in solving fixed-time trajectory optimization problems with fixed initial and final states across multiple systems.
The goal is to understand how each method's solve time scales with increasing system dimension.
In each of these trials input and joint limits are enforced, but there are no obstacle avoidance constraints.
The objective function is formulated as minimizing control effort along the trajectory.

\begin{figure}[t]
    \centering
    \includegraphics[width=1\linewidth]{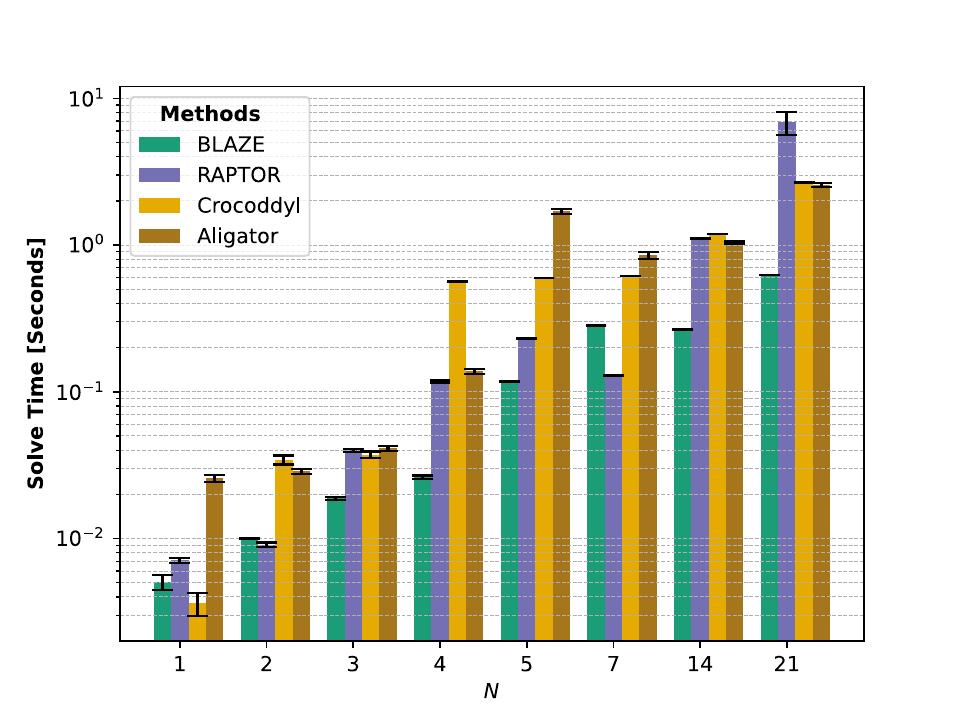}
    \caption{
    A bar plot comparing the mean solve times for four different trajectory optimization algorithms: \methodnamenew{}, RAPTOR, Crocoddyl and Aligator.
    For $N \leq 5 $ the results correspond to the 1-5 link pendulum, $N=7$ corresponds to the Kinova Gen3 Arm, $N=14$ corresponds to a bimanual system of two Kinova Gen3 Arms, $N=21$ corresponds to a trimanual system of three Kinova Gen3 Arms.
    Each experiment was run ten times.
    Overall, we see that \methodnamenew{} shows better scalability and solve times than the other methods as the system dimension increases.
    }
    \label{fig:scalability-t-solve-no-obstacles}
\end{figure}

We do this evaluation for a 1-, 2-, 3-, 4-, 5-link pendulum model, a 7DOF Kinova Gen3 arm, a double 7DOF Kinova Gen3 arm and a triple 7DOF Kinova Gen3.
Figure \ref{fig:scalability-t-solve-no-obstacles} shows how each of the aforementioned methods scale with increasing $N$.
We see that \methodnamenew{} is able to scale more favorably that the other methods and solves faster than all the other methods for most of the evaluated robot systems with $N>1$.
Refer to tables \ref{table:blaze_parameters_experiment_1}, \ref{table: parameters aligator experiment 1} and \ref{table: crocoddyl scalability} for the specific parameter values used for \methodnamenew{}, Aligator and Crocoddyl respectively.

\subsection{Kinova Gen3 Sphere Obstacle Avoidance}
This section compares \methodnamenew{} and Aligator in solving multiple fixed-time trajectory optimizations to get the Kinova Gen3 from some initial state to some final state while avoiding sphere obstacles.
Note, we do not run Crocoddyl or RAPTOR in these experiments, because they did not have any examples doing sphere obstacle avoidance.
In these experiments, we design $\Xinit$ such that it starts in collision, to evaluate how effective the proposed method is in moving the initial guess out of constraint violation during Phase 1 before solving Phase 2.
Each of these methods is ran to convergence at a small tolernace, ensuring fully converged solutions.
We evaluate each of these methods on 20 different scenarios where either the obstacles are randomly placed and $\Xinit$ is generated to be in collision.
For 10 of these scenarios, there are 5 obstacles, while for the other 10 there are 10 obstacles.
The objective function is formulated to minimize control effort along the trajectory.
For each method, we gave a 60s time budget to be able to generate a trajectory.
Being unable to generate the trajectory within that time counted as an unsuccessful trial.

As illustrated in Table \ref{table: kinova obstacles}, \methodnamenew{} is able to generate trajectories much faster than Aligator when ran to full convergence, and is able to generate trajectories that do not collide with obstacles for all the scenarios.
To explore how Aligator fared when allowed to return suboptimal solutions we rerun the same experiment over the same scenarios but allow Aligator to return its suboptimal solutions after each iteration. 
Table \ref{table: kinova obstacles iterations} shows that \methodnamenew{} fully converges in less time than Aligator’s first iteration and achieves a lower objective cost even compared to Aligator’s third iteration.

\begin{table}[t] 
    \centering
    \begin{tabular}{ |c | c | c | }
        \hline
        \textbf{Method Name} & \textbf{\methodnamenew{}} & \textbf{Aligator} \\
        \hline
        \textbf{Success Rate [\%]} & 100 & 65 \\
        \hline
        \textbf{Objective Cost} & 738.7 ± 157.1 & 635.9 ± 108.81 \\
        \hline
        \textbf{Solve Time [s]} & 0.75 ± 0.27 & 27 ± 11.06 \\
        \hline
    \end{tabular}
    \caption{Comparison of Success Rate, Objective Cost and Solve Time for Kinova Gen3 trajectory optimization experiments with obstacles.
    Note that the Objective Cost and Solve Time numbers are just presented for scenarios where both methods were successful.
   }
    \label{table: kinova obstacles}
\end{table}

\begin{table}[h] 
\vspace{-0.2cm}
    \centering
    \begin{tabular}{ |c | c | c | c |}
        \hline
        \textbf{Method Name} & \textbf{\methodnamenew{}} & \textbf{Aligator} it $=1$ &  \textbf{Aligator} it $=3$ \\
        \hline
        \textbf{Objective Cost} & 733.2 ± 166.3 & 884.8 ± 140.4 & 775.3 ± 120.5 \\
        \hline
        \textbf{Solve Time [s]} & 0.84 ± 0.33 & 0.91 ± 0.85 & 1.45 ± 0.84 \\
        \hline
    \end{tabular}
    \caption{\footnotesize Objective Cost and Solve Time for \methodnamenew{} compared with Aligator after its 1st and 3rd iterations on Kinova Gen3 obstacle-avoidance tasks, where Aligator is permitted to return intermediate suboptimal solutions.
    Results are aggregated over all 20 scenarios, rather than only successful trials as in Table \ref{table: kinova obstacles} due to a large proportion of the Aligator optimizations after the first iteration not satisfying the success criteria.  
    }
    \vspace{-0.5cm}
    \label{table: kinova obstacles iterations}
\end{table}

\subsection{Kinova Gen3 Hard Task-Based Scenarios}

This section compares \methodnamenew{}, Aligator, and RAPTOR under a similar setup as the previous subsection, with two key differences. First, obstacles are now cuboids arranged to resemble more task-oriented scenarios for manipulators that would be deployed in domestic and industrial settings.
These scenarios involve actions such as reaching into shelves, retracting arms out of shelves and under shelves, and placing items in confined bins, reflecting real-world manipulation tasks.
Second, the time limit for each method is extended to 120s to accommodate scenarios with a greater number of obstacles.
For all of these experiments each method must generate a 2-second long trajectory, which is a relatively short time to execute many of these scenarios.
This time constraint forces the methods to produce highly dynamic solutions while simultaneously satisfying state and input constraints, as well as obstacle avoidance, making the scenarios particularly challenging.
A trial is considered unsuccessful if a method fails to generate a trajectory within the allotted time or does not satisfy the earlier success criteria.

\begin{figure}[h]
    \centering
     \includegraphics[trim={0cm, 0cm, 0cm, 2cm},clip,width=1.0\linewidth]{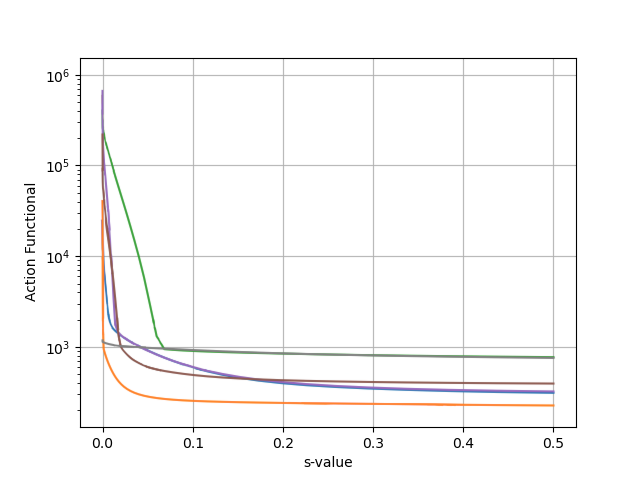}
    \caption{
    \footnotesize A plot showing the evolution of the Action Functional versus the AGHF evolution parameter $s$ for all the hard task–based scenario experiments.}
    \label{fig: convergence plots}
    \vspace{-0.0cm}
\end{figure}

Figure \ref{fig: convergence plots} shows the evolution of the Action Functional versus the AGHF evolution parameter $s$ for all the hard task–based scenario experiments. 
Each curve corresponds to one scenario and exhibits the Action functional decreasing and fully converging to a steady‐state value.

\begin{figure}[t]
    \centering
    \centering
    \begin{subfigure}{0.49\textwidth}
        \includegraphics[trim={5cm, 0cm, 5cm, 0cm},clip,width=1\columnwidth]{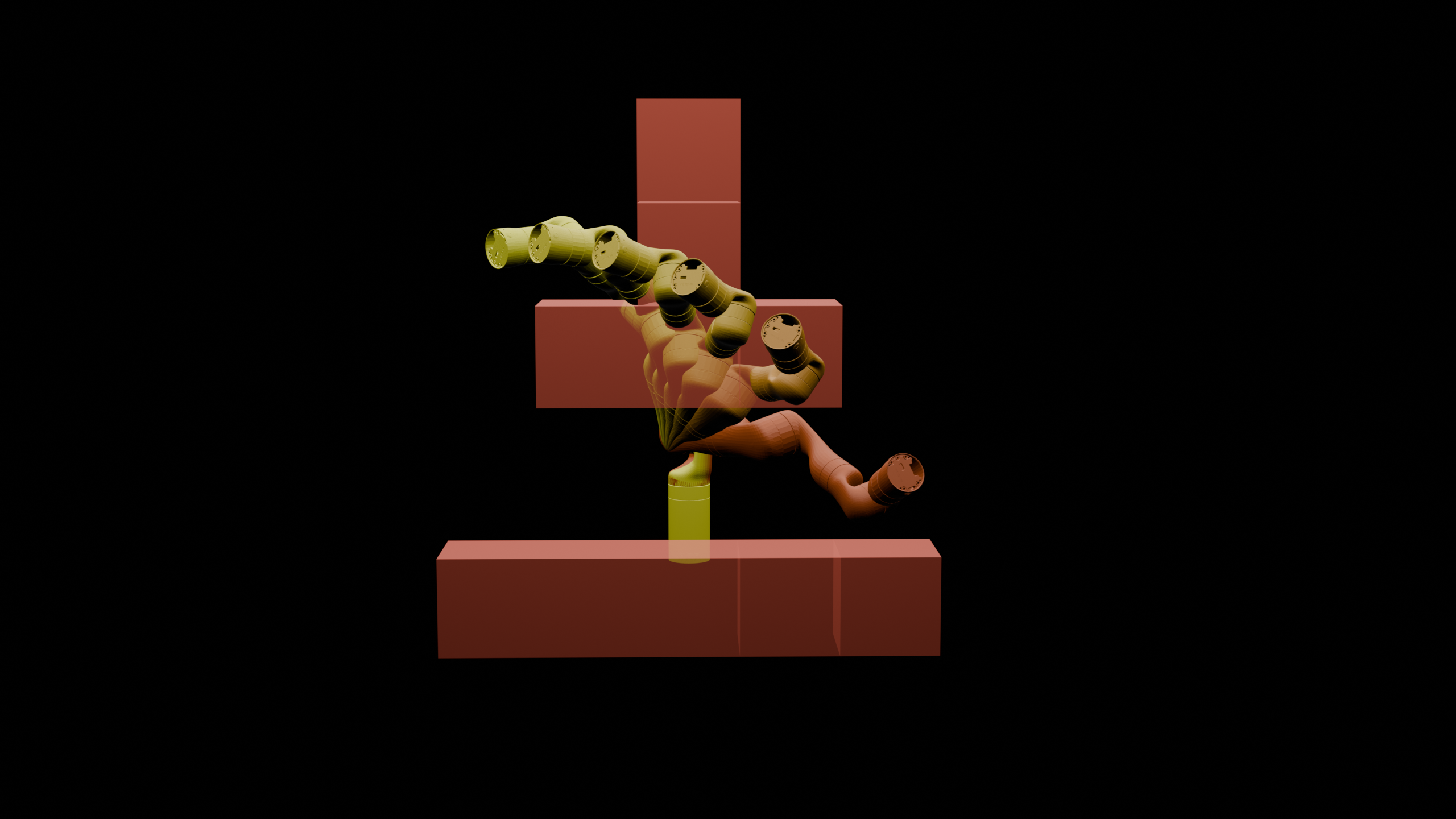}
    \end{subfigure}
    \hfill
    \vspace*{0.1mm}
    \begin{subfigure}{0.49\textwidth}
        \includegraphics[trim={5cm, 0cm, 5cm, 0cm},clip,width=1\columnwidth]{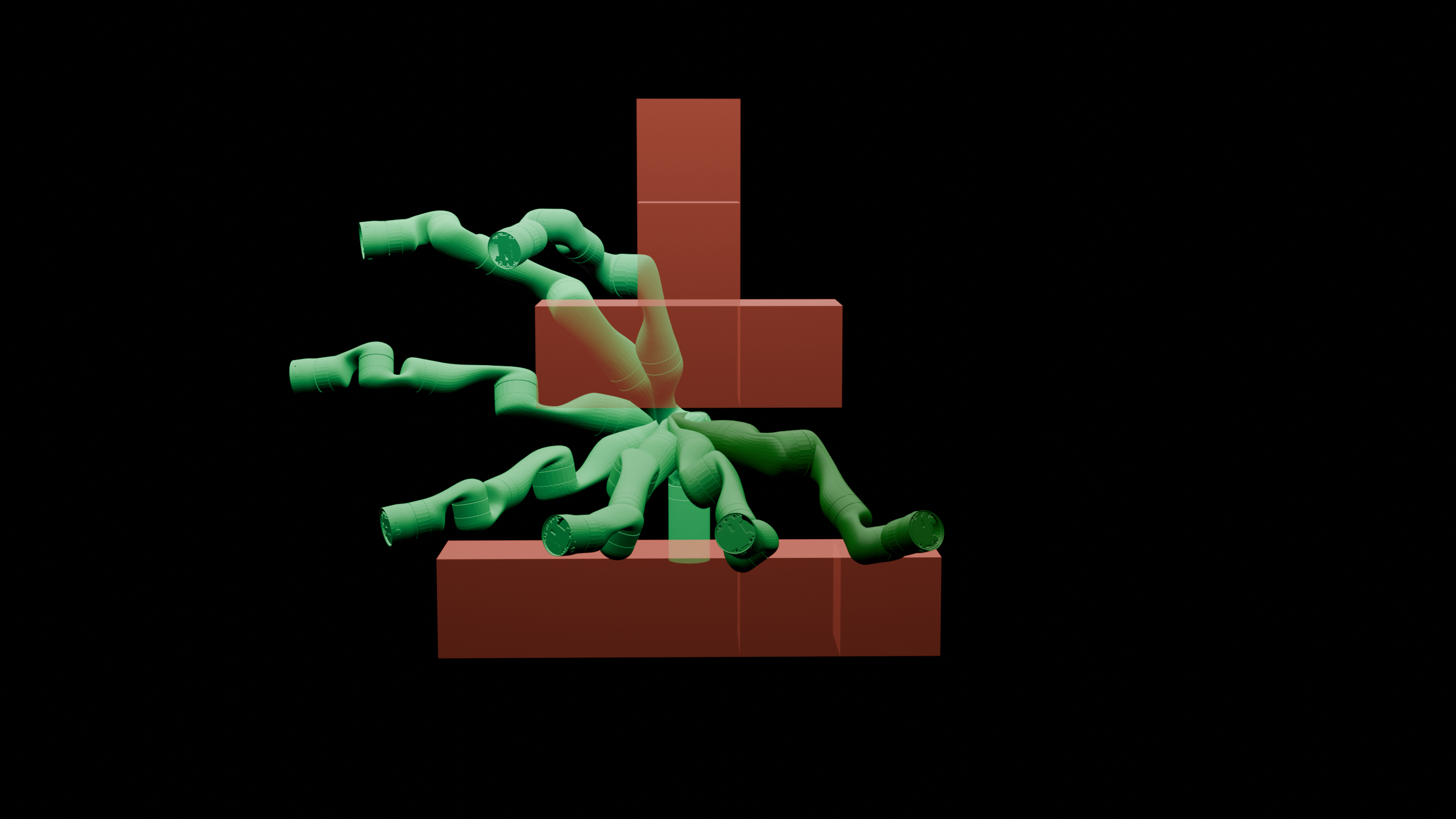}
    \end{subfigure}
    \caption{
    This figure shows a visualization of one of the task-based scenarios (scenario 4).
    In each subfigure, the color gradient illustrates the evolution in time where the start configuration is in the darkest shade and the end configuration in the lightest.
    The initial trajectory for the scenario is shown in the top image in yellow and collides with the obstacles along the path.
    The proposed algorithm is able to push this initial guess out of collision and generate the optimal collision-free solution shown in green in 2.19s for this scenario, whereas the other comparison methods cannot find a solution within the allotted time for this scenario.
    }
    \label{fig:blender_realistic_scenario}
\end{figure}

Table \ref{table: detailed experiment 3} summarizes the performance of the three optimal control algorithms for the 6 different realistic scenarios.
In this table ``F" denotes that a scenario was considered  a failure as per the success criteria introduced in Section \ref{subsec: success}.
\methodnamenew{} demonstrates a higher success rate, completing 100\% of trials compared to 33.3\% for Aligator and 0\% for RAPTOR.
Figure \ref{fig:blender_realistic_scenario} illustrates a solution generated by \methodnamenew{} for one of the task-based scenarios.

\begin{table}[h] 
    \centering
    \begin{tabular}{|c|c|c|}
        \hline
        \multirow{2}{*}{\textbf{Scenario}} & \textbf{Objective} & \textbf{Solve} \\
        & \textbf{Cost} & \textbf{Time [s]} \\
        \hline
        \multirow{3}{*}{1} & \textbf{\textcolor{blaze_color}{723.02}} & \textbf{\textcolor{blaze_color}{0.86 ± 0.01}}  \\ 
        \cline{2-3}
        & \textbf{\textcolor{aligator_color}{F}} & \textbf{\textcolor{aligator_color}{F}} \\
        \cline{2-3}
        & \textbf{\textcolor{raptor_color}{F}} & \textbf{\textcolor{raptor_color}{F}} \\
        \hline
        \multirow{3}{*}{2} & \textbf{\textcolor{blaze_color}{773.05}} & \textbf{\textcolor{blaze_color}{2.45 ± 0.02}}\\ 
        \cline{2-3}
        & \textbf{\textcolor{aligator_color}{291.35}} & \textbf{\textcolor{aligator_color}{57.21 ± 0.13}} \\
        \cline{2-3}
        & \textbf{\textcolor{raptor_color}{F}} & \textbf{\textcolor{raptor_color}{F}} \\
        \hline
        \multirow{3}{*}{3} & \textbf{\textcolor{blaze_color}{983.82}} & \textbf{\textcolor{blaze_color}{1.6 ± 0.01}} \\ 
        \cline{2-3}
        & \textbf{\textcolor{aligator_color}{F}} & \textbf{\textcolor{aligator_color}{F}} \\
        \cline{2-3}
        & \textbf{\textcolor{raptor_color}{F}} & \textbf{\textcolor{raptor_color}{F}} \\
        \hline
        \multirow{3}{*}{4} & \textbf{\textcolor{blaze_color}{1147.68}} & \textbf{\textcolor{blaze_color}{2.19 ± 0.01}} \\
        \cline{2-3}
        & \textbf{\textcolor{aligator_color}{F}} & \textbf{\textcolor{aligator_color}{F}} \\
        \cline{2-3}
        & \textbf{\textcolor{raptor_color}{F}} & \textbf{\textcolor{raptor_color}{F}} \\
        \hline
        \multirow{3}{*}{5} & \textbf{\textcolor{blaze_color}{223.78}} & \textbf{\textcolor{blaze_color}{1.28 ± 0.01}} \\ 
        \cline{2-3}
        & \textbf{\textcolor{aligator_color}{187.68}} & \textbf{\textcolor{aligator_color}{68.46 ± 0.19}} \\
        \cline{2-3}
        & \textbf{\textcolor{raptor_color}{F}} & \textbf{\textcolor{raptor_color}{F}} \\
        \hline
        \multirow{3}{*}{6} & \textbf{\textcolor{blaze_color}{496.95}} & \textbf{\textcolor{blaze_color}{3.1 ± 0.02}} \\ 
        \cline{2-3}
        & \textbf{\textcolor{aligator_color}{F}} & \textbf{\textcolor{aligator_color}{F}} \\
        \cline{2-3}
        & \textbf{\textcolor{raptor_color}{F}} & \textbf{\textcolor{raptor_color}{F}} \\
        \hline      
    \end{tabular}
    \caption{Detailed comparison of Objective Cost and Solve Time for the different task-based scenarios. 
    All the problems consider input, state and cuboid obstacle constraints.
    \methodnamenew{} is depicted in \textcolor{blaze_color}{\textbf{green}}, Aligator is depicted in \textcolor{aligator_color}{\textbf{gold}}, and RAPTOR is depicted in \textcolor{raptor_color}{\textbf{purple}}. In this table "F" denotes that a scenario was considered a failure as per the success criteria introduced in Section \ref{subsec: success}.
    }
    \label{table: detailed experiment 3}
\end{table}

\subsection{Hardware Experiments}
We demonstrate the performance of \methodnamenew{} on hardware on the Kinova Gen3 robot on a number of challenging obstacle avoidance scenarios, where the robot must navigate from some initial position to some final position while adhering to the robot's state and input constraints.
Here, the dynamic feasibility of \methodnamenew{} as well as its ability to generate trajectories that satisfy the real hardware constraints is critical to enabling the robot to successfully execute the generated trajectory.
On the robot, we leverage a passivity-based controller to track the trajectories we generate.

\begin{figure}[t]
    \centering
    \begin{subfigure}{0.49\linewidth}
        \centering
        \includegraphics[width=\linewidth]{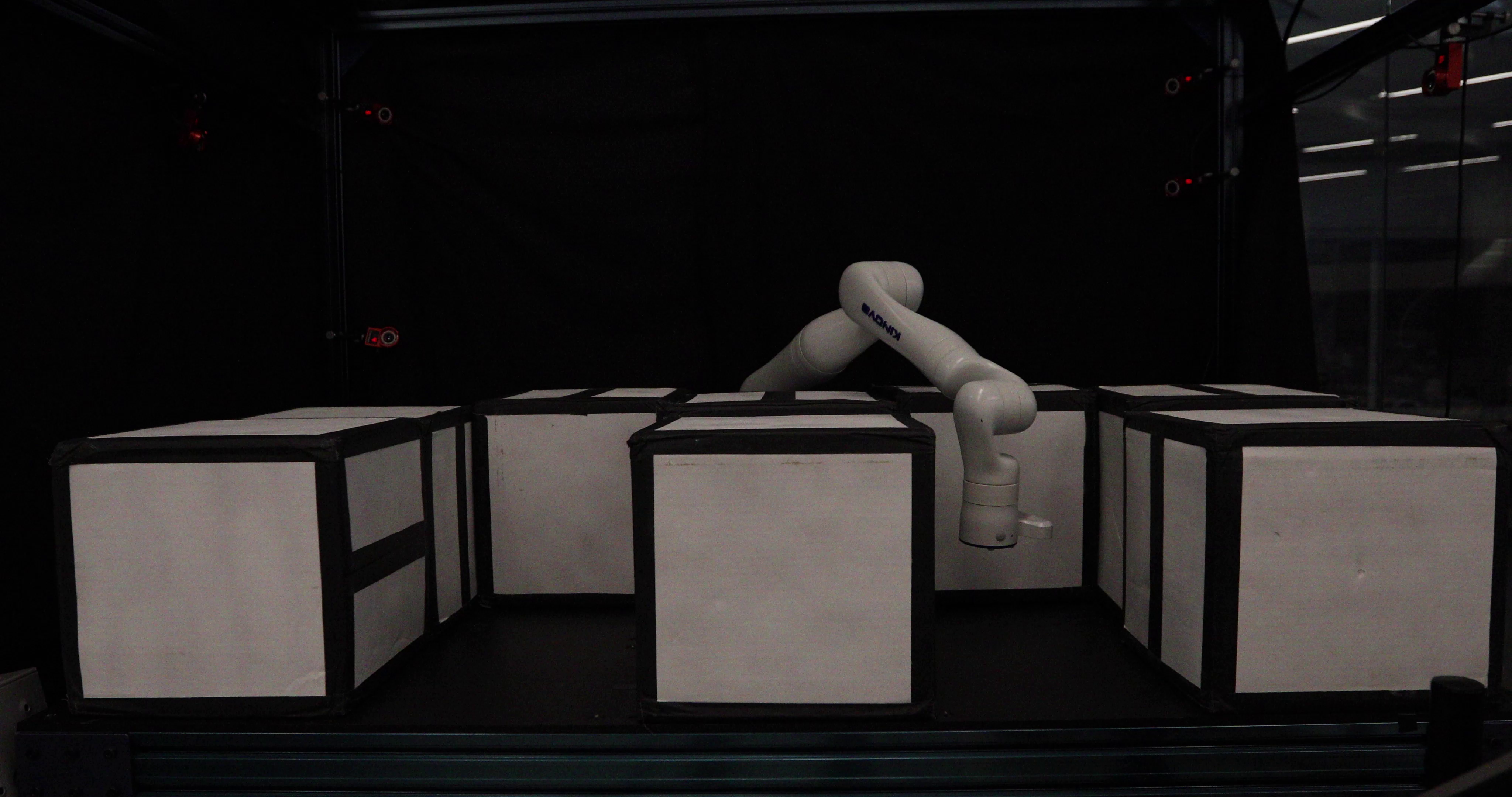}
        \caption{Still frame at the initial time $t=0$s.}
    \end{subfigure}
    \begin{subfigure}{0.49\linewidth}
        \centering
        \includegraphics[width=\linewidth]{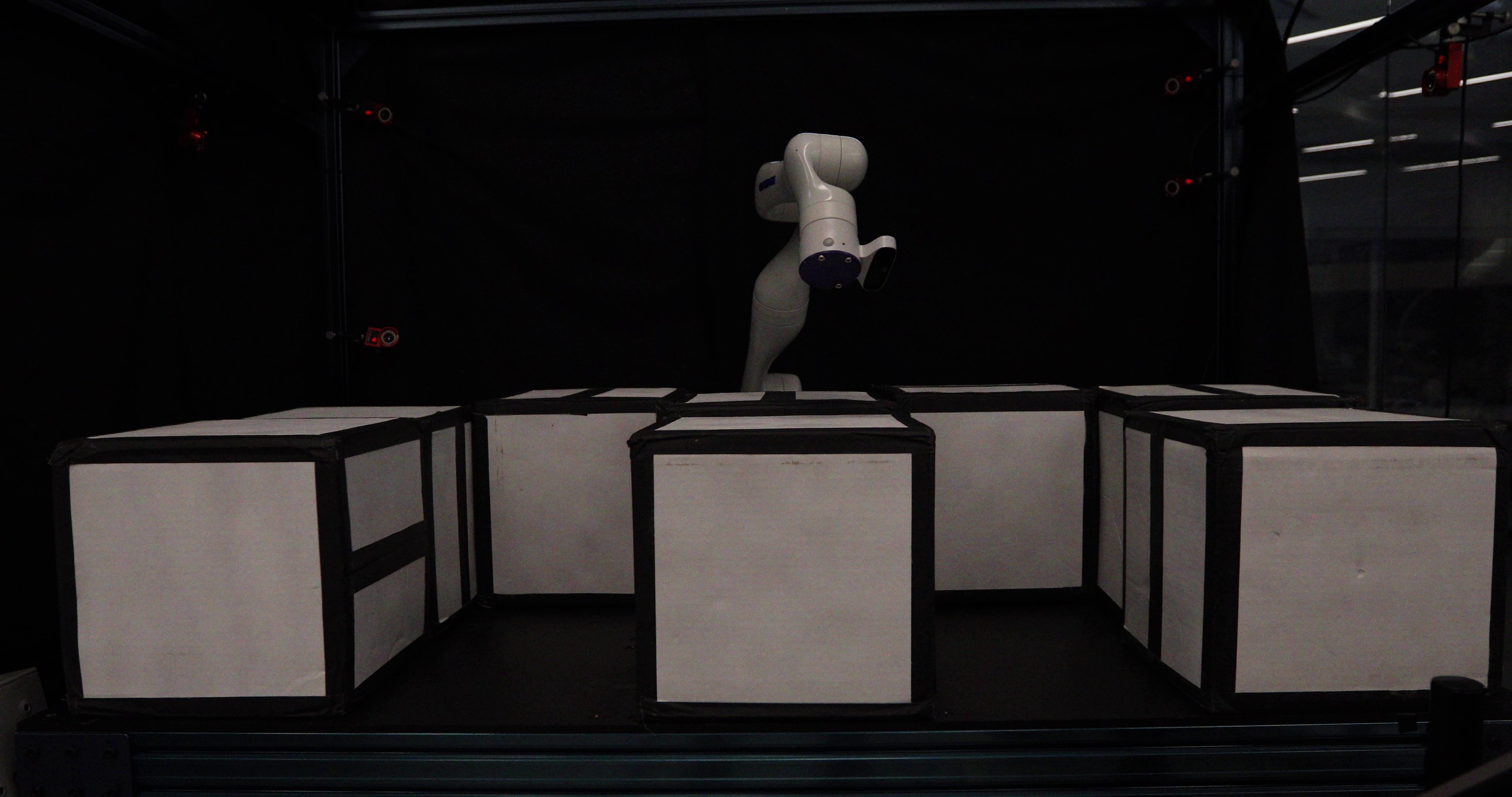}
        \caption{Still frame at time $t=0.67$s.}
    \end{subfigure} \\
    \begin{subfigure}{0.49\linewidth}
        \centering
        \includegraphics[width=\linewidth]{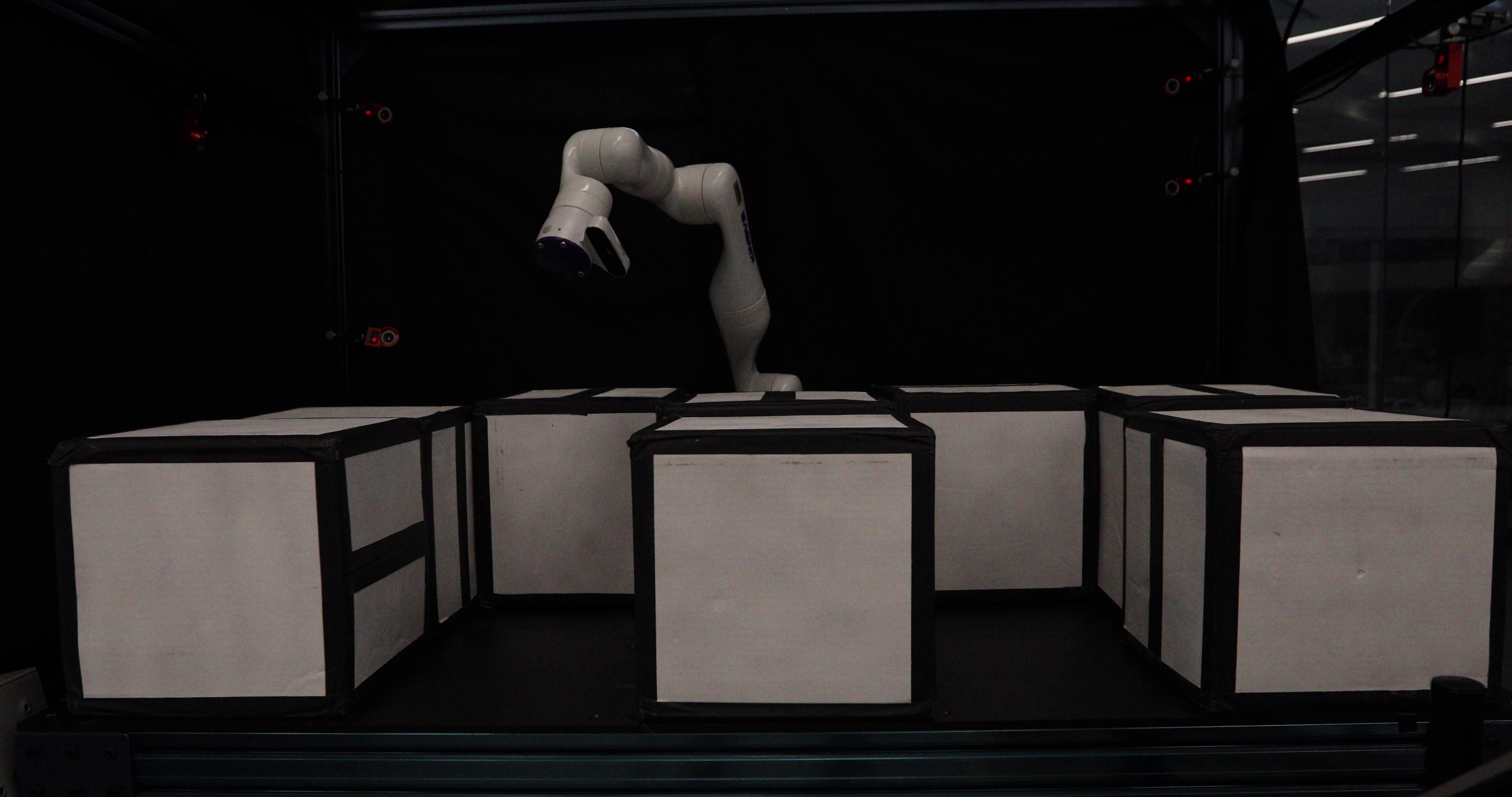}
        \caption{Still frame at time $t=1.34$s.}
    \end{subfigure}
    \begin{subfigure}{0.49\linewidth}
        \centering
        \includegraphics[width=\linewidth]{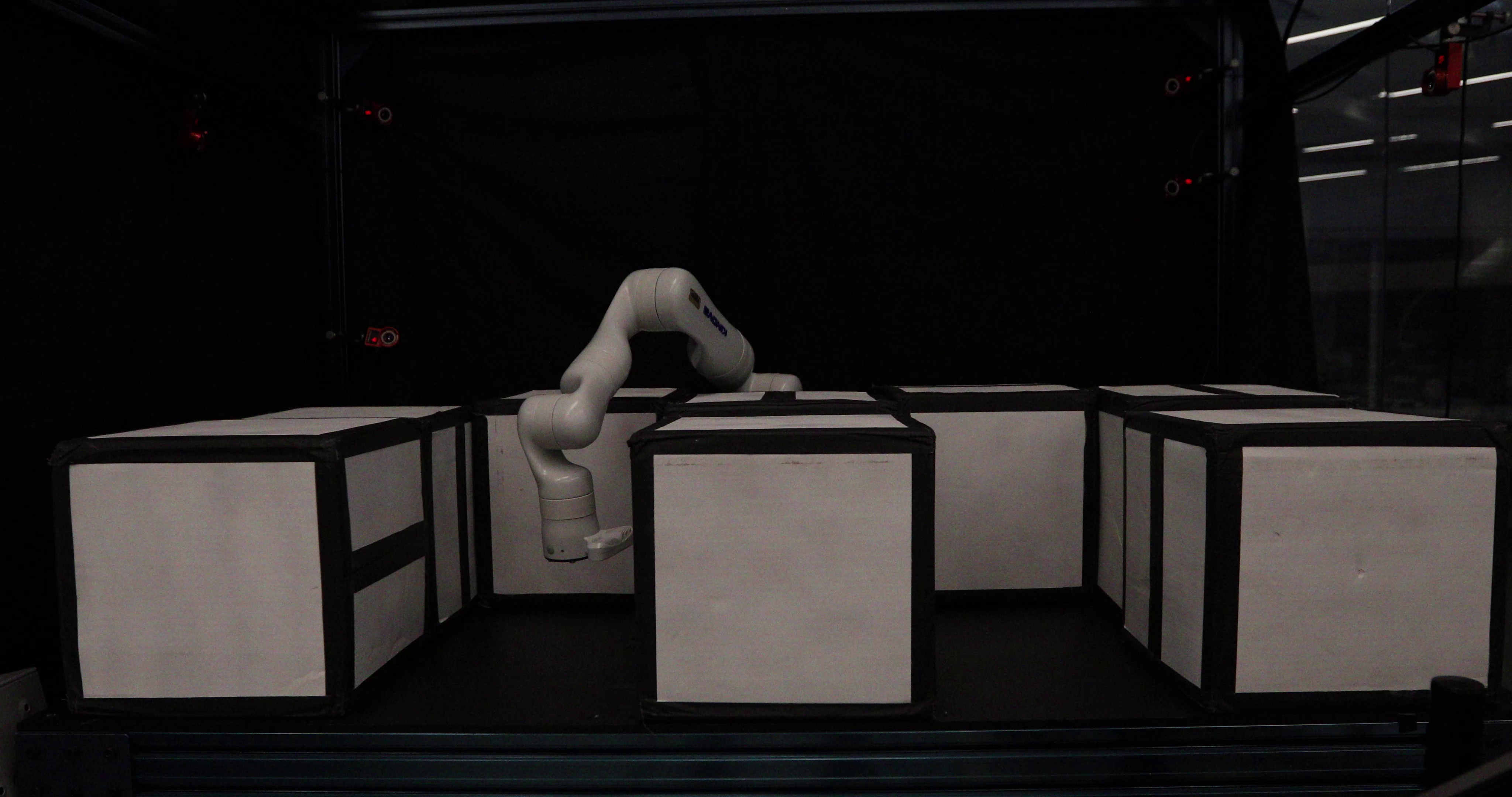}
        \caption{Still frame at the final time $t=2.0$s.}
    \end{subfigure}
    \caption{
        This figure shows a series of still frames of the Kinova arm following the trajectory generated by \methodnamenew{} for Scenario 2 in Table \ref{table: detailed experiment 3}.  
        Each subfigure depicts the arm at a different point in time.  
        The initial guess collides with the obstacle.  
        The proposed algorithm is able to push this initial guess out of collision and generate the optimal collision-free solution in 2.45s, whereas Aligator does it in 57.21s.
    }
    \label{fig: scene 3 in hardware}
\end{figure}

The project page includes videos showcasing various challenging scenarios, such as a rapid pull-and-place maneuver where the robot retracts its arm from a simulated shelf and swiftly places it in a bin (hardware scenario 1, solved in 3.1 seconds); a fast retraction from a confined space while avoiding obstacles (hardware scenario 2, solved in 2.74 seconds); and a precise pick-and-place trajectory requiring navigation through tight grasp points (hardware scenario 3, solved in 2.45 seconds, as shown in Figure \ref{fig: scene 3 in hardware}).
Each of these trajectories is executed within a strict 2-second duration, requiring the robot to move dynamically while adhering to state and input constraints. 
These experiments highlight \methodnamenew{}'s ability to rapidly generate dynamically feasible trajectories that fully account for the robot's full-order dynamics, demonstrating its effectiveness in real-world task execution. 

\section{Conclusion}
\label{sec:conclusion}

This work proposes \methodnamenew{}, a generalized AGHF-based trajectory optimization framework that generalizes the AGHF PDE methods to arbitrary cost functions, significantly expanding the range of feasible trajectories.
By leveraging the AGHF PDE, this method evolves an initial, potentially infeasible trajectory into a dynamically feasible solution while optimizing a specified objective function, allowing for rapid trajectory generation.
A key contribution of this work is the Phase1-Phase2 framework, made possible by the generalized cost function formulation.
This Phase1-Phase2 framework addresses a limitation of previous AGHF approaches -- requiring constraint-satisfying initial guesses to ensure fast convergence.
Phase1 leverages the generalized cost function to drive the trajectory into the feasible set before Phase2 optimizes the objective while maintaining feasibility.
This allows the initial guess to violate constraints while ensuring rapid convergence to a valid solution. 
Additionally, a new method for enforcing input constraints within AGHF is introduced, enabling trajectory optimization with realistic actuation limits without increasing the state dimension, as was required in previous AGHF-based approaches.
Simulations and hardware experiments demonstrate that \methodnamenew{} can generate trajectories for high-dimensional robotic systems under multiple constraints, with constraint-violating initial guesses faster than state-of-the-art trajectory optimization methods.
\section{Limitations}
\label{sec:limitations}

\methodnamenew{} has its limitations: similar to other AGHF formulations because the action functional \eqref{eqn:action functional k} incorporates dynamic constraints through a penalty term, the AGHF solution does not minimize the objective function as dramatically as other optimal control methods. 
In particular, traditional optimal control methods may construct lower cost trajectories; however, in practice, because generated trajectories satisfy all constraints (including input constraints) this may not be too restrictive.
In fact if one's objective is to generate trajectories rapidly that are dynamically feasible, \methodnamenew{} may be the right approach.
One other difficulty with \methodnamenew{} stems from its requirement to balance between the constraint penalty, $k_{cons}$, and the dynamic feasibility penalty, $k$, which can require some tuning.

\section{Acknowledgements}
\label{sec: acknowledgements}

This work was supported by AFOSR MURI
FA9550-23-1-0400, and the Automotive Research Center (ARC).
The authors thank Adam Li for his support and for supplying the software tools that enabled the creation of some of the visuals in this work.

\bibliographystyle{IEEEtranN}
\bibliography{references}

\pagebreak
\clearpage
\appendices
\section{Proof of Lemma \ref{lem:AGHF convergence}}
\label{sec: appendix aghf dissipation}
\begin{proof}
    To prove this result, we construct the variation of the Action Function with respect to $s$ using Taylor Expansion and Integration by Parts. 
    This enables us to leverage the definition of the AGHF PDE to prove the desired result. 

    Consider the variation of \eqref{eqn:action functional}:
     \begin{align}
    \label{eqn:act functional variation}
        \Act(\X_{ s + \delta}) = \int_0^T  L(\X(t, s + \delta), \Xd(t, s + \delta)) dt.
    \end{align}
    The Taylor Expansion of the integrand of \eqref{eqn:action functional} is:
    \begin{align}
    \label{eqn:taylor expansion lagrangian}
    \begin{split}
        L(\X(t, s + \delta), \Xd(t, s + \delta)) = \Lxsxds + \delta \dldxs \dxsds\\
        & \hspace{-4.5cm}  + \delta \dldxds \frac{\partial \Xd_s}{\partial s} + o(\delta)
    \end{split}
    \end{align}
    where for convenience we have dropped the arguments in the terms on the right hand side. 
    Substituting \eqref{eqn:taylor expansion lagrangian} into \eqref{eqn:act functional variation}
    \begin{align}
        \label{eqn: act functional variation expanded}
        \begin{split}
            \Act(\X_{ s + \delta}) = \Act(\X_{ s }) + \delta \int_0^T \Bigg( \dldxs \dxsds  \\
            & \hspace{-5.2cm} + \dldxds \dxdsds + o(\delta) \Bigg) dt.
        \end{split}
    \end{align}
    Integrating  \eqref{eqn: act functional variation expanded} by parts and applying the boundary conditions \eqref{eqn: bcs aghf}, we obtain:
    \begin{align}
    \begin{split}
        \Act(\X_{ s + \delta})
        & = \Act(\X_{ s }) + \delta \int_0^T \frac{\partial \X_s}{\partial s} \bigg( 
            \dldxs(\X_s(t), \dot{\X}_s(t)) \notag \\
        & - \ddt \dldxds(\X_s(t), \dot{\X}_s(t)) 
        \bigg) \,  dt + o(\delta).
    \end{split}
    \end{align}
    Rearranging the terms, we obtain
    \begin{align}
    \begin{split}
        \frac{ \Act(\X_{ s + \delta})  - \Act(\X_{ s}) } {\delta} = \int_0^T \frac{\partial \X_s}{\partial s} \bigg( 
            \dldxs(\X_s(t), \dot{\X}_s(t)) \\
        & \hspace{-4.55cm} - \ddt \dldxds(\X_s(t), \dot{\X}_s(t)) 
        \bigg) \, dt + \frac{o(\delta)}{\delta}
    \end{split}
    \end{align}

    Treating $\Delta s = \delta$, we take $\Delta s \rightarrow 0$ to obtain
    \begin{align}
        \label{eqn:s derivative act functional prev}
        \frac{\partial \Act}{\partial s} = \int_0^T \frac{\partial \X_s}{\partial s} \bigg( 
            \dldxs(\X_s(t), \dot{\X}_s(t)) \notag \\
         - \ddt \dldxds(\X_s(t), \dot{\X}_s(t)) 
        \bigg) \, dt
    \end{align}
    Substituting \eqref{eqn:AGHF} into \eqref{eqn:s derivative act functional prev}
    \begin{align}
        \frac{\partial \Act}{\partial s} =  - \int_0^T \frac{\partial \X_s^T}{\partial s} M \frac{\partial \X_s}{\partial s}.
    \end{align}
     Thus we see that
     \begin{align}
         \frac{\partial \Act}{\partial s} \leq 0 \iff M \succeq 0.
     \end{align}
\end{proof}
\section{Proof of Theorem \ref{thm:conv_dyn_feasible} }
\label{sec: appendix aghf convergence}
Before proving this result, we prove the following result that is used in the proof of the result:
\begin{lem}
\label{lem:helper_lemma}
Suppose the Cost Function in the definition of the Action Functional \eqref{eqn:action functional k} is coercive with constants $\alpha$ and $\beta$.
Then for any $C \in \R$, we have $\left\{ \X \in C^1([0,T] \to \R^{2N}) \mid \Act(\X) \leq C \right\} \subseteq \left\{ \X \in C^1([0,T] \to \R^{2N}) \mid \|\dot{\X}\|^2_{L^\infty} + \|\X\|^2_{L^\infty} \leq \frac{\beta+C T}{\alpha} \right\}$
\end{lem}
\begin{proof}
We prove this result by contradiction.
We first assume that $\X$ or $\dot{X}$ are not essentially bounded.
By utilizing the definition of coercivity, we can lower bound the integrand with something that grows arbitrarily large which leads to a contradiction. 

Consider $\X \in\left\{ \X \in C^1([0,T] \to \R^{2N}) \mid \Act(\X) \leq C \right\}$ such that either $\X \notin L^\infty$ or $\dot{\X}\notin L^\infty$.
Without loss of generality assume that $\X \notin L^\infty$.
This implies that for every large $M \geq 0$ there is a set of positive measure $I \subset [0,T]$, where $\| \X \|_{L^{\infty}} \geq M$, which implies that through coercivity that 
\begin{equation}
 \kd  \| \dot{\X}_{P1}(t) - \X_{P2}(t)\|_2^2 +  c(\X(t),\dot{\X}(t),u(t)) \geq \alpha M - \beta,
\end{equation}
for $t \in I$.
Integrating the Lagrangian over $I$ gives:
\begin{multline}
    \int_I \left(\kd  \| \dot{\X}_{P1}(t) - \X_{P2}(t)\|_2^2 +  c(\X(t),\dot{\X}(t),u(t)) \right) dt\\
    \geq \left(\alpha M - \beta\right) \lambda(I),
\end{multline}
where $\lambda(I) > 0$ describes the size of $I$. 
Because $M$ can be made arbitrarily large, this leads to a contradiction. 
To construct the bound, note that the largest violation would occur if $I=T$. 
Using this observation with the previous argument provides the desired result.

\end{proof}

Now we use the previous lemma to prove Theorem \ref{thm:conv_dyn_feasible}: 
\begin{proof}

To prove this result, we begin by defining a sublevel set of the Action Functional that we prove is invariant under the AGHF Flow.
Next, we compute the norm of the error between a trajectory belonging to this invariant set and the integrated trajectory. 
By applying the Bellman-Gronwall Inequality and the fact that follows from Lemma \ref{lem:helper_lemma} we are able to prove our result. 

Because the path planning problem is feasible, there exists $\X^*$ that satisfies the dynamics \eqref{eq: control affine dyn}. 
Plug $\X^*$ into the Action Functional and pick some $C_1$ such that $C_1 > \Act(\X^*)$. 
Note $C_1$ is independent of $k_d$. 
Let $X =\{ \X\in AC([0,T] \to \mathbb{R}^{2N}) \mid \X(0) = \x_0, \X(T) = \x_f\}$ where $AC$ denotes the space of absolutely continuous functions.
Let $\Omega^{AC}_{k_d} = \{ \X\in X \mid \Act(\X) < C_1\}$ which is not empty because it at least contains $\X^*$. 
Because $\Act$ is continuous over $X$ with respect to the norm on absolutely continuous functions, $\Omega^{AC}_{k_d}$ is open. 
Because $C^1([0,T] \to \mathbb{R}^{2N})$ is dense in $AC([0,T] \to \mathbb{R}^{2N})$, $\Omega_{k_d}^{'}:= \Omega^{AC}_{k_d} \cap C^1([0,T] \to \mathbb{R}^{2N})$ is open as well.
From Lemma \ref{lem:AGHF convergence}, we know that $\Act$ is non-increasing so $\Omega^{'}_{k_d}$ is invariant. 
Let $\Omega_{k_d}$ be the region of attraction to $\Omega^{'}_{k_d}$; that is $\Omega^{'}_{k_d} \subset \Omega_{k_d} \subset C^1([0,T] \to \mathbb{R}^{2N})$ and all AGHF solutions $x_s$ with an initial condition starting from $\Omega_{k_d}$ will converge to the invariant set $\Omega^{'}_{k_d}$ when $s$ increases.
Consequently, when $s_{max}$ is sufficiently large, $\Act(\X_{s_{max}}) \leq C_1$ and $\X_{s_{max}}$ is continuous. 

Before moving on to compute the error trajectory, we make one observation. 

Define the curve $\X(t) = \X_{s_{max}}(t)$ and $\ubar(t) \in  \R^{2N}$ be given by:
\begin{align}
    \label{eqn:fake control}
    \ubar(t) = 
    \begin{bmatrix}
        \uu_c(t) \\
        \uu(t)
    \end{bmatrix} = \bar{F}(\X(t))^{-1} (\dot{\X}(t) - \Fd(\X(t))).
\end{align}
Substituting \eqref{eqn:fake control} into \eqref{eqn:action functional k}
\begin{align}
    C_1 \geq \int_0^T \left( \kd \|\uu_c(t) \|_2^2 +  c(\X(t), \Xd(t), \uu(t)) \right) dt.
\end{align}
From which we conclude
\begin{align}
    \label{eqn:bound on uc}
    \int_0^T \|\uu_c(t) \|_2^2 dt \leq \frac{C_1}{\kd}.
\end{align}

Next, define the dynamics of the integrated system:
\begin{align}
    \begin{split}
    \tilde{x}(t) &= \Fd(\tilde{x}(t)) + F(\tilde{x}(t))\uu(t), \\
    \tilde{x}(0) &= \x_0.
    \end{split}
\end{align}
Define the error $e(t) = \X(t) - \tilde{x}(t)$ and then the error dynamics are
\begin{align}
\begin{split}
    \dot{e}(t) &= \Xd(t) - \xdint(t) \\
    &= (\Fd(\X(t)) - \Fd(\xint(t))) \\
    &\quad + (F(\X(t)) - F(\xint(t))) \uu(t) \\
    &\quad + \Fc(\X(t))\uu_c(t).
\end{split}
\end{align}
Integrating both sides
\begin{align}
\begin{split}
   \|e(t)\|_2^2 &= \int_0^t \bigg( (\Fd(\X(\tau)) - \Fd(\xint(\tau))) \\
    &\quad + (F(\X(\tau)) - F(\xint(\tau))) \uu(\tau) \\
    &\quad + \Fc(\X(\tau))\uu_c(\tau) \bigg) d\tau.
\end{split}
\end{align}
Squaring the norm on both sides
\begin{align}
\begin{split}
    \|e(t)\|_2^2 &= \int_0^t  \| (\Fd(\X(\tau)) - \Fd(\xint(\tau))) \\
    &\quad + (F(\X(\tau)) - F(\xint(\tau))) \uu(\tau) \\
    &\quad + \Fc(\X(\tau))\uu_c(\tau) \|_2^2 d\tau.   
\end{split}
\end{align}
Applying the Cauchy-Schawrtz Inequality
\begin{align}
\begin{split}
    \|e(t)\|_2^2 &\leq t \int_0^t  \| (\Fd(\X(\tau)) - \Fd(\xint(\tau))) \\
    &\quad + (F(\X(\tau)) - F(\xint(\tau))) \uu(\tau) \\
    &\quad + \Fc(\X(\tau))\uu_c(\tau) \|_2^2 d\tau.   
\end{split}
\end{align}
Applying the Triangle Inequality
\begin{align}
\begin{split}
    \|e(t)\|_2^2 &\leq t \int_0^t \bigg( \| \Fd(\X(\tau)) - \Fd(\xint(\tau))\|  \\
    &\quad +  \|(F(\X(\tau)) - F(\xint(\tau))) \uu(\tau) \| \\
    &\quad + \| \Fc(\X(\tau))\uu_c(\tau) \| \bigg)^2 d\tau.
\end{split}
\end{align}
Using the Power Mean Inequality
\begin{align}
\begin{split}
    \|e(t)\|_2^2 &\leq 3t \int_0^t \bigg( \| \Fd(\X(\tau)) - \Fd(\xint(\tau))\|_2^2  \\
    &\quad +  \|(F(\X(\tau)) - F(\xint(\tau))) \uu(\tau) \|_2^2 \\
    &\quad + \| \Fc(\X(\tau))\uu_c(\tau) \|_2^2 \bigg) d\tau.
\end{split}
\end{align}

Because $F_c$ is continuous and $\X$ is bounded in the $L^\infty$ sense because of Lemma \ref{lem:helper_lemma}, we know there exists a $C_2 \geq 0$ such that almost everywhere:
\begin{equation}
\| \Fc(\X(\tau))\uu_c(\tau) \|_2 \leq \| \Fc(\X(\tau))\|_2 \| \uu_c(\tau) \|_2 \leq C_2 \| \uu_c(\tau) \|_2. 
\end{equation}
Using this observation, the fact that $F_d$ and $F$ are globally Lipschitz, and applying \eqref{eqn:bound on uc}:
\begin{align}
\begin{split}
    \|e(t)\|_2^2 &\leq 3t \int_0^t (Y_1^2 + Y_2^2 \|\uu(\tau)\|_2^2)\|e(t)\|_2^2 d\tau \\
        &\quad + \frac{3t C_1 C_2^2}{k_d}
\end{split}
\end{align}
Finally, we can apply the Bellman-Grönwall Inequality:
\begin{align}
\label{eqn:error inequality belabbas}
\begin{split}
    \|e(t)\|_2^2 &\leq \bigg(  \frac{3t C_1 C_2^2}{k_d} \bigg) \\
    &\quad \exp \bigg( 3t \int_0^t (Y_1^2 + Y_2^2 \|\uu(\tau)\|_2^2 ) d\tau \bigg)
\end{split}
\end{align}

To bound the extracted control input $u$, recall its Definition \ref{defn:control_extraction} and note that all functions that define the operation are continuous. 
Because of the boundedness of $\X$ in the $L^\infty$ sense which follows from Lemma \ref{lem:helper_lemma} one can prove that there exists a $C_3$ such that
\begin{align}
    \label{eqn:bound on u}
    \int_0^T \|\uu(\tau) \|_2^2 d\tau \leq C_3
\end{align}
Substituting the inequalities \eqref{eqn:bound on uc} and \eqref{eqn:bound on u} into \eqref{eqn:error inequality belabbas} we conclude that:
\begin{align}
\|\xint(t) - \X(t)\|_2 \leq \sqrt{ \frac{3tC_1 C_2^2}{k_d} \exp{\bigg( 3t (Y_1^2 t + Y_2^2 C_3) \bigg)} }
\end{align}
and by substituting $t=T$ we obtain the desired conclusion.
\end{proof}

\section{Proof of Lemma \ref{lem: control inverse dyn} }
\label{sec: appendix control RNEA}
\begin{proof}

Recall from \eqref{eqn:control extraction} that $\uu_s(t)$ is given by:
\begin{equation}
\label{eqn:control extraction appendix}
    \uu_s(t) = \begin{bmatrix} 0_{N \times  N} & I_{N \times N} \end{bmatrix}
    \bar{F}(\X_s(t))^{-1} (\dot{\X}_s(t) - \Fd(\X_s(t)))
\end{equation}
We begin by substituting \eqref{eqn: Fd} and \eqref{eqn: F} into \eqref{eqn:control extraction appendix} 
\begin{align}
\begin{split}
    \uu_s(t) = \begin{bmatrix} 0_{N \times  N} & I_{N \times N} \end{bmatrix} \cdot \\
    & \hspace{-3.6cm} \cdot \underbrace{\begin{bmatrix}
        I_{N \times (2N-m)} & 0_{N \times m}\\
        0_{N \times (2N-m)} & H(\xpone(s, t)) \end{bmatrix}^T }_{(\bar{F}(\X_s(t))^{-1})^T} \cdot \\   
    & \hspace{-3cm} \underbrace{
    \begin{bmatrix}
    \xdpone - \xptwo \\
    \xdptwo + \Hinv(\xpone(s, t)) C(\xpone(s, t), \xptwo(,s))
    \end{bmatrix}
    }_{\Xd - \Fd}   
\end{split}
\end{align}
Expanding and simplifying yields
\begin{align}
\begin{split}
    \uu_s(t) = H(\xpone(s, t) \xdptwo(s, t) + C(\xpone(s, t), \xptwo(s, t)).
\end{split}
\end{align}
Note that this equation corresponds exactly to the robot dynamics presented in \eqref{eq: manipulator dyn} when $B=I$.

\end{proof}
\section{Experiment hyperparameters}
\label{appendix: experiments parameters}
As pointed out in \cite{adu2024bringheatrapidtrajectory}, Aligator and Crocoddyl have several parameters that affect the performance. 
The parameter that is common to both methods and affects the solve time most is the time discretization $\Delta t$.
The methods also allow the user to choose a running cost through the choice of weights $w_u$, $w_x$. 
The running cost is the 2-norm of the input summed to the 2-norm of the state with weights $w_u$ and $w_x$ respectively. 
The methods allow the user to weigh a terminal cost through the weight $w_{xf}$.
Crocoddyl allows penalizing state inequality constraint by a factor $w_{state}$ and the terminal value by $w_{state, f}$.
Finally, Aligator also allows one to specify $\epsilon_{tol}$ and $\mu_{init}$ which control the convergence properties of the optimization.

Tables \ref{table: parameters aligator experiment 1}, \ref{table: parameters aligator experiment 2}, \ref{table: parameters for aligator experiment 3} summarizes the parameters used by Aligator in the scalability experiment, the sphere obstacle avoidance experiment, and the realistic obstacle avoidance scenario respectively.
Table \ref{table: crocoddyl scalability} describes the parameters used by Crocoddyl for the scalability experiment.

RAPTOR optimizes over Bezier polynomial coefficients representing the state trajectory. The primary hyperparameters for RAPTOR include the Bezier polynomial degree $p$ and the number of points $N$ where constraints are evaluated along the trajectory. Additionally, since RAPTOR utilizes IPOPT as its optimizer, several IPOPT-specific hyperparameters can be tuned, such as \texttt{linear\_solver}, which influences the speed of Newton steps, and \texttt{mu\_strategy}, which affects convergence speed and stability.
For both the swing-up experiments and the obstacle avoidance experiments, we used the same parameter set across all scenarios, following the recommendations provided in RAPTOR’s official GitHub repository.
Table \ref{table:parameters_raptor} details these parameters.

\begin{table}[t]
    \centering
    \renewcommand{\arraystretch}{1.2}
    \begin{tabular}{ | l | c | }
        \hline
        \textbf{Parameter}  & \textbf{Value} \\
        \hline
        Bezier Polynomial Degree (\( p \)) & 7 \\ \hline
        Constraint Evaluation Points (\( N \)) & 20 \\ \hline
        IPOPT mu Strategy (\texttt{mu\_strategy}) & adaptive \\ \hline
        IPOPT Linear Solver (\texttt{linear\_solver}) & ma57 \\
        \hline
    \end{tabular}
    \caption{Parameter settings for RAPTOR in the swing-up experiments.}
    \label{table:parameters_raptor}
\end{table}

\methodnamenew{} implements a pseudospectral MOL algorithm to solve the AGHF, which is similar to the approach adopted by \cite{adu2024bringheatrapidtrajectory}.
As a result, we refer to \cite{adu2024bringheatrapidtrajectory} for a complete exposition of the typical parameters of the AGHF. 
In this paper we extend the parameters of the constrains $k_{cons}$ and $c_{cons}$ to ones designed per constraint.
For instance, $c_{cuboid}$ corresponds to $c_{cons}$ for the cuboid constraints. 
Similarly, $k_{sph}$ corresponds to the weight on the constraint that encodes collision avoidance from sphere obstacles.
The parameters we choose for each experiment for \methodnamenew{} are summarized in Tables \ref{table:blaze_parameters_experiment_1}, \ref{table:blaze_parameters_experiment_2}, and \ref{table:blaze_parameters_experiment_3}.

\begin{table}[h]
    \centering
    \renewcommand{\arraystretch}{1.2}
    \begin{tabular}{|l |c | c | c | c | c | c|}
    \hline
        DOF  & $w_u$  & $w_x$  & $w_{xf}$  & $\Delta t$  & $\epsilon_{tol}$  & $\mu_{\text{init}}$ \\
    \hline 
        1  & $10^{-4}$  & $10^{-3}$  & $10^{-4}$  & $10^{-2}$  & $10^{-2}$  & $10^{-8}$   \\
        \hline
        2   & $10^{-4}$  & $10^{-2}$  & $10^{-2}$  & $10^{-2}$  & $10^{-2}$  & $10^{-7}$   \\
        \hline
         3   & $10^{-3}$  & $1$        & $1$        & $10^{-2}$  & $10^{-2}$  & $10^{-7}$   \\
         \hline
        4   & $10^{-3}$  & $1$        & $10^{-4}$  & $10^{-2}$  & $10^{-2}$  & $10^{-7}$   \\
        \hline
         5   & $10^{-3}$  & $1$        & $10^{-6}$  & $10^{-2}$  & $10^{-2}$  & $10^{-7}$   \\
         \hline
        7  & $10^{-2}$  & $1$        & $1$        & $10^{-2}$  & $10^{-3}$  & $10^{-7}$   \\
        \hline
        14 & $10^{-2}$  & $1$        & $1$        & $10^{-2}$  & $10^{-3}$  & $10^{-7}$   \\
        \hline
        21 & $10^{-2}$  & $1$      & $1$        & $10^{-2}$  & $10^{-3}$  & $10^{-7}$   \\
    \hline
    \end{tabular}
    \caption{Parameter settings for Aligator for the Scalability Experiment}
    \label{table: parameters aligator experiment 1}
\end{table}

\begin{table}[h]
    \centering
    \renewcommand{\arraystretch}{1.2}
    \begin{tabular}{ | l | c | c | c | c | c | c | c}
        \hline
        DOF  & \textbf{$w_u$}  & \textbf{$w_x$}  & \textbf{$w_{xf}$}  & \textbf{$\Delta t$}  & $\epsilon_{tol}$  & $\mu_{\text{init}}$ \\
        \hline
        7 & $10^{-2}$  & $1$  & $1$  & $10^{-2}$  & $10^{-3}$  & $10^{-7}$ \\
        \hline
    \end{tabular}
    \caption{Parameter settings for Aligator for the Sphere Obstacle Avoidance experiment}
    \label{table: parameters aligator experiment 2}
\end{table}

\begin{table}[h]
    \centering
    \renewcommand{\arraystretch}{1.2}
    \begin{tabular}{| l | c | c | c | c | c | c | c |}
        \hline
        DOF  & \textbf{$w_u$}  & \textbf{$w_x$}  & \textbf{$w_{xf}$}  & \textbf{$\Delta t$}  & $\epsilon_{tol}$  & $\mu_{\text{init}}$   \\
        \hline
        7 & $10^{-2}$  & $1$  & $1$  & $10^{-2}$  & $10^{-3}$  & $10^{-4}$ \\
        \hline
    \end{tabular}
    \caption{Parameter settings for Aligator for the realistic scenario experiment}
    \label{table: parameters for aligator experiment 3}
\end{table}

\begin{table}[H]
    \centering
    \renewcommand{\arraystretch}{1.2}
    \begin{tabular}{| l | c | c | c | c | c | c |}
        \hline
        DOF & $w_u$ & $w_x$ & $w_{xf}$ & $w_{\text{state}}$ & $w_{\text{state},f}$ & $\Delta t$ \\
        \hline
        1   & $10^{-3}$ & $10^{-2}$ & $10$   & $10$   & $10$   & $10^{-2}$ \\
        \hline
        2   & $10^{-4}$ & $10^{-4}$ & $10^3$ & $10$   & $10$   & $10^{-2}$ \\
        \hline
        3   & $10^{-4}$ & $10^{-3}$ & $1$    & $10$   & $10$   & $10^{-2}$ \\
        \hline
        4   & $10^{-3}$ & $10^{-2}$ & $10$   & $10$   & $10$   & $10^{-3}$ \\
        \hline
        5   & $10^{-4}$ & $10^{-2}$ & $10$   & $10$   & $10$   & $10^{-3}$ \\
        \hline
        7   & $10^{-4}$ & $10^{-2}$ & $10^2$ & $10$   & $10^{-1}$ & $10^{-3}$ \\
        \hline
        14  & $10^{-3}$ & $10^{-2}$ & $10^2$ & $1$    & $10^{-1}$ & $10^{-3}$ \\
        \hline
        21  & $10^{-3}$ & $10^{-2}$ & $10^2$ & $1$    & $10^{-1}$ & $10^{-3}$ \\
        \hline
    \end{tabular}
    \caption{Parameter settings for Crocoddyl for Scalability experiment}
    \label{table: crocoddyl scalability}
\end{table}

\begin{table}[t]
    \vspace{-100pt}
    \centering
    \begin{minipage}{\linewidth}
        \centering
        \renewcommand{\arraystretch}{1.2}
    \begin{tabular}{| l | c | c | c | c | c | c |}
        \hline
        DOF & k  & $c_{\text{state}}$  & $k_{\text{state}}$  & $c_{\text{input}}$  & $k_{\text{input}}$  & $s_{\max}$ \\
        \hline
        1   & $10^9$  & 50  & $10^5$  & 50  & $10^5$  & 0.01 \\
        \hline
        2   & $10^{10}$ & 50  & $10^5$  & 50  & $10^5$  & 0.01 \\
        \hline
        3   & $10^6$  & 50  & $10^{10}$ & 50  & $10^{10}$ & 1 \\
        \hline
        4   & $10^6$  & 50  & $10^7$  & 50  & $10^7$  & 1 \\
        \hline
        5   & $10^6$  & 1   & $10^5$  & 1   & $10^5$  & 1 \\
        \hline
        7   & $10^6$  & 200 & $10^9$  & 200 & $10^9$  & 100 \\
        \hline
        14  & $10^{14}$ & 200 & $10^6$  & 200 & $10^6$  & 10 \\
        \hline
        21  & $10^{14}$ & 200 & $10^6$  & 200 & $10^6$  & 10 \\
        \hline
    \end{tabular}
    \caption{Parameter settings for \methodnamenew{} Scalability Experiment}
    \label{table:blaze_parameters_experiment_1}
    \end{minipage}
\end{table}

\begin{table}[h!]
    \vspace{-270pt}
    \centering
    \begin{minipage}{\linewidth}
        \centering
    \renewcommand{\arraystretch}{1.2}
    \begin{tabular}{| l | c | c |}
        \hline
        Parameter  & Phase 1  & Phase 2 \\
        \hline
        $s_{\max}$        & 10       & 10       \\
        \hline
        $k$               & $10^4$   & $10^4$   \\
        \hline
        $p$               & 7        & 7        \\
        \hline
        $c_{\text{sph}}$  & 100      & 100      \\
        \hline
        $c_{\text{state}}$ & 100     & 100      \\
        \hline
        $c_{\text{input}}$ & 100     & 100      \\
        \hline
        $k_{\text{sph}}$  & $10^7$   & $10^7$   \\
        \hline
        $k_{\text{state}}$ & $10^6$   & $10^6$   \\
        \hline
        $k_{\text{input}}$ & $10^6$   & $10^6$   \\
        \hline
    \end{tabular}
    \caption{Parameter settings for \methodnamenew{} for Sphere Obstacle avoidance experiment.}
    \label{table:blaze_parameters_experiment_2}
    \end{minipage}
\end{table}

\begin{table}
\vspace{-270pt}
    \centering
    \renewcommand{\arraystretch}{1.2}
    \begin{tabular}{| l | c | c |}
        \hline
        Parameter  & Phase 1  & Phase 2 \\
        \hline
        $s_{\max}$        & 0.5     & 0.5     \\
        \hline
        $k$               & $10^4$  & $10^4$  \\
        \hline
        $p$               & 12      & 12      \\
        \hline
        $c_{\text{sph}}$  & 200     & -       \\
        \hline
        $c_{\text{cuboid}}$  & -       & 200     \\
        \hline
        $c_{\text{state}}$ & 200     & 200     \\
        \hline
        $c_{\text{input}}$ & 200     & 200     \\
        \hline
        $k_{\text{sph}}$  & $10^5$  & -       \\
        \hline
        $k_{\text{cuboid}}$  & -       & $10^5$  \\
        \hline
        $k_{\text{state}}$ & $10^4$  & $10^4$  \\
        \hline
        $k_{\text{input}}$ & $10^4$  & $10^4$  \\
        \hline
    \end{tabular}
    \caption{Parameter settings for \methodnamenew{} for realistic scenario experiment.}
    \label{table:blaze_parameters_experiment_3}
\end{table}

\end{document}